\documentclass[10pt,twocolumn,letterpaper]{article}
\usepackage{makecell}
\usepackage{colortbl}
\usepackage[table]{xcolor}
\usepackage{multirow}
\usepackage{tabularx}
\usepackage{subfigure}
\usepackage{color,xcolor}

\usepackage{listings}
\usepackage{xcolor}

\usepackage{cvpr}              
\usepackage{array}
\definecolor{cvprblue}{rgb}{0.21,0.49,0.74}
\usepackage[pagebackref,breaklinks,colorlinks,allcolors=cvprblue]{hyperref}

\definecolor{blue}{RGB}{136, 185, 251}
\definecolor{red}{RGB}{247, 154, 154}
\def\paperID{43020} 
\def\confName{CVPR}
\def\confYear{2026}

\title{Ultrasound-CLIP: Semantic-Aware Contrastive Pre-training for Ultrasound Image-Text Understanding}

\author{Jiayun Jin$^1$, Haolong Chai$^1$, Xueying Huang$^1$, Xiaoqing Guo$^2$, Zengwei Zheng$^1$, Zhan Zhou$^3$, \\Junmei Wang$^4$, Xinyu Wang$^5$, Jie Liu$^{6,*}$, and Binbin Zhou$^{1,}$\thanks{Corresponding authors: Binbin Zhou (\href{mailto:bbzhou@hzcu.edu.cn}{bbzhou@hzcu.edu.cn}) and Jie Liu (\href{mailto:jliu.ee@my.cityu.edu.hk}{jliu.ee@my.cityu.edu.hk})}\\[0.5mm]
$^1$Hangzhou City University~~~~$^2$Hong Kong Baptist University~~~~$^3$Zhejiang University \\ $^4$Women's Hospital, School of Medicine, Zhejiang University \\ $^5$The First Affiliated Hospital, Zhejiang University School of Medicine \\ $^6$City University of Hong Kong \\ [0.5mm]
{\small Project:~\href{https://github.com/ZJUDataIntelligence/Ultrasound-CLIP}{https://github.com/ZJUDataIntelligence/Ultrasound-CLIP}}\\ {\small Data:~\href{https://huggingface.co/datasets/JJY-0823/US-365K}{https://huggingface.co/datasets/JJY-0823/US-365K}}
}

\begin{document}
\maketitle
\begin{abstract}
Ultrasound imaging is widely used in clinical diagnostics due to its real-time capability and radiation-free nature. However, existing vision-language pre-training models, such as CLIP, are primarily designed for other modalities, and are difficult to directly apply to ultrasound data, which exhibit heterogeneous anatomical structures and diverse diagnostic attributes.
To bridge this gap, we construct US-365K, a large-scale ultrasound image–text dataset containing 365k paired samples across 52 anatomical categories. We establish Ultrasonographic Diagnostic Taxonomy (UDT) containing two hierarchical knowledge frameworks. Ultrasonographic Hierarchical Anatomical Taxonomy standardizes anatomical organization, and Ultrasonographic Diagnostic Attribute Framework formalizes nine diagnostic dimensions, including body system, organ, diagnosis, shape, margins, echogenicity, internal characteristics, posterior acoustic phenomena, and vascularity.
Building upon these foundations, we propose Ultrasound-CLIP, a semantic-aware contrastive learning framework that introduces semantic soft labels and semantic loss to refine sample discrimination. Moreover, we construct a heterogeneous graph modality derived from UDAF's textual representations, enabling structured reasoning over lesion–attribute relations.
Extensive experiments with patient-level data splitting demonstrate that our approach achieves state-of-the-art performance on classification and retrieval benchmarks, while also delivering strong generalization to zero-shot, linear probing, and fine-tuning tasks. 

\end{abstract}   
\section{Introduction}
\label{sec:intro}

Ultrasound imaging has emerged as the second most widely deployed medical imaging modality \cite{gehealthcare2024autonomous}, valued for its real-time, non-invasive, and radiation-free visualization of internal organs \cite{shen2021artificial, le2025u2}. It serves as the first-line diagnostic tool across diverse clinical scenarios, from prenatal \cite{maani2025fetalclip} to thyroid screening \cite{cao2023artificial}. However, ultrasound imaging presents unique challenges that distinguish it from other medical imaging modalities. The fundamental physics of ultrasound relies on tissue-specific acoustic properties \cite{harvey2002advances}, where different tissues exhibit distinct echogenic patterns based on their acoustic impedance. This results in significant appearance variations of identical pathologies across anatomical regions, creating a notoriously steep learning curve that requires years of specialized training to achieve proficiency.

\begin{figure}
    \centering
    \includegraphics[width=\linewidth]{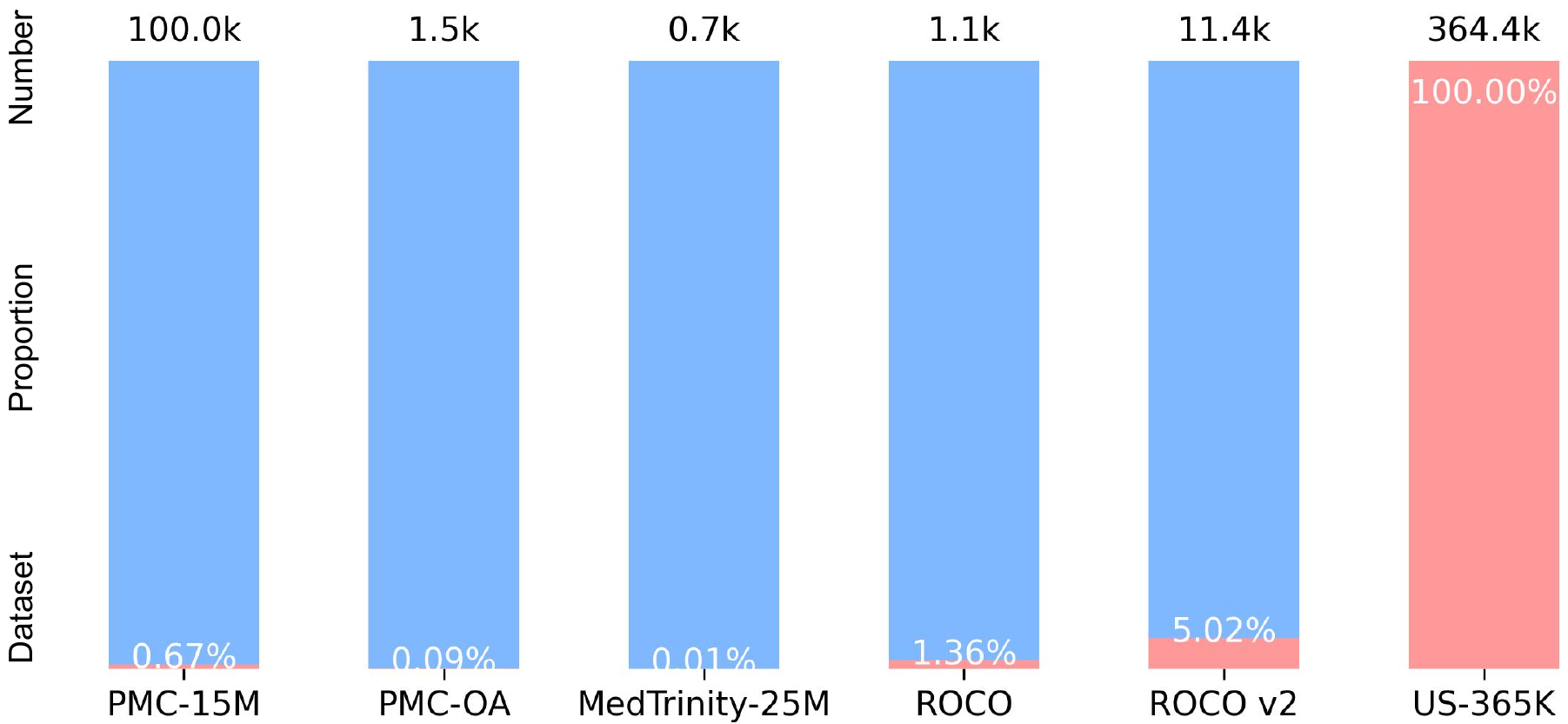}
    \caption{\textbf{Ultrasound image statistics  across major benchmarks.} The \textcolor{red}{Red} segment and internal percentage show the proportion of ultrasound images, while the \textcolor{blue}{blue} segment show the remaining modalities. The top label indicates the absolute number (in thousands). Our US-365K is the first large-scale, 100\% dedicated ultrasound dataset.}
    \label{fig1}
\vspace{-0.2in}
\end{figure}

Recent developments in vision-language pre-training (VLP) models \cite{wan2023med,yang2022vision,chen2023vlp,wu2024mmclip} have shown significant promise in supporting physicians' medical image understanding, potentially mitigating the steep learning curve associated with ultrasound interpretation. 
Building upon the success of Contrastive Language-Image Pre-training (CLIP) \cite{radford2021learning}, some works leverage medical literature and radiology reports to construct medical vision-language datasets, such as ROCO \cite{pelka2018roco}, MedTrinity-25M \cite{xie2024medtrinity}, PMC-OA \cite{lin2023pmc}, PMC-15M \cite{zhang2023biomedclip} and UniMed \cite{khattak2024unimed}. However, our analysis of existing datasets reveals a striking imbalance: \textit{ultrasound data remains severely underrepresented, accounting for less than 5\% of existing medical VLP datasets, despite its clinical prevalence} (see Figure ~\ref{fig1}). 

This oversight underscores the urgent need for a dedicated and comprehensive ultrasound vision-language dataset. 
Building such a dataset is non-trivial due to the inherent characteristics of ultrasound. 
Unlike the comprehensive views of CT and MRI, ultrasound imaging is localized and captures operator-dependent planes. More importantly, ultrasound reports utilize ultrasound-specific attribute to describe findings, such as diagnosis, echogenicity, and posterior acoustic phenomena and vascularity, which are entirely foreign to models trained on general radiology data. 
To address these challenges, we introduce the \textbf{Ultrasonographic Diagnostic Taxonomy (UDT)}. The UDT integrates \textbf{Ultrasonographic Hierarchical Anatomical Taxonomy (UHAT)}, which standardizes anatomical organization, and \textbf{Ultrasonographic Diagnostic Attribute Framework (UDAF)}, which formalizes nine key diagnostic dimensions. 
Grounded in the UDT, we construct \textbf{US-365K}, the first large-scale and semantic-rich dataset for the ultrasound understanding, comprising 365K image-text pairs. 

Building upon this US-365K dataset, we propose \textbf{Ultrasound-CLIP}, a semantic-aware contrastive framework designed specifically to overcome critical failures of existing models when applied to sonography. 
Previous CLIP-based frameworks \cite{eslami2023pubmedclip,wang2022medclip,khattak2024unimed,lin2023pmc,zhang2023biomedclip,zou2025alignment,ko2025bringing} are not optimal for the complexities of ultrasound data due to two reasons. First, they struggle with semantic ambiguity. Ultrasound reports contain diverse and unstructured diagnostic expressions, making it difficult for standard contrastive learning to define valid positive and negative pairs in such a semantically dense environment. Second, they lack domain-specific structural priors to model complex relationships among diagnostic attributes, which are essential for clinical interpretation. Hence, Ultrasound-CLIP is committed to address the two failures. It introduces a \textbf{UDAF-guided heterogeneous graph encoder} to explicitly model intra-text dependencies between diagnostic labels and their attributes and further leverages the \textbf{UDAF taxonomy to generate semantic soft labels} with continuous similarity metric that recognizes fine-grained semantic overlap. A dual-branch contrastive learning objective is then utilized to learn high-level representations for downstream tasks. 


%


In summary, our contributions are fourfold:
\begin{itemize}
    \item We introduce the UDT, a novel hierarchical knowledge framework to formalize and unify ultrasound-specific diagnostic semantics.
    \item We construct US-365K, the first large-scale, semantically-rich VLP benchmark dataset for the ultrasound modality, grounded in our UDT.
    \item We propose Ultrasound-CLIP, a modality-tailored contrastive framework that integrates semantic alignment and heterogeneous graph encoding to address semantic ambiguity and model structured clinical reasoning.
    \item Our framework set a new state-of-the-art, demonstrating a significant improvement over the best-performing medical CLIP baseline and robust generalization. All code and dataset are publicly available.
\end{itemize}

\section{Related works}
\label{sec:relatedwork}

\subsection{Medical Vision-Language Datasets}

High-quality, large-scale datasets are crucial for developing robust medical vision-language models. 
Recent large-scale medical VLP datasets, such as PMC-15M \cite{zhang2023biomedclip}, PMC-OA \cite{lin2023pmc}, MedTrinity-25M \cite{xie2024medtrinity}, ROCO \cite{pelka2018roco} and ROCO v2 \cite{ruckert2024rocov2}, have enabled significant progress. 
However, these datasets are overwhelmingly dominated by radiology (CT, MRI) and pathology. Our analysis confirms a stark underrepresentation of ultrasound: it accounts for less than 0.67\% PMC-15M, 0.09\% PMC-OA, a mere 0.01\% of MedTrinity-25M, 1.36\% of ROCO, and 5.02\% of ROCO v2 (see Figure~\ref{fig1}). 
Recent efforts have constructed valuable datasets for specific sonographic domains, such as Fetal-CLIP \cite{maani2025fetalclip}, ReMUD \cite{wang2025haiburemudreasoningmultimodalultrasound}, and EchoPrime \cite{nature2025com}. While these datasets are powerful for their intended tasks, their specialized focus on narrow anatomical regions highlights the need for a comprehensive, pan-anatomy VLP benchmark for general ultrasound. To our knowledge, our work, US-365K, is the first to address this critical resource gap, providing broad coverage across diverse anatomical systems.



\subsection{Medical Contrastive Language-Image Pre-training}
CLIP \cite{radford2021learning} and its variants \cite{ICLR2024_d1450d6c,zhai2023sigmoid,liu2023clip,liu2024universal,khattak2024unimed,du2024ret,LAC2025IJCAI} have revolutionized multimodal understanding by learning joint representations from image-text pairs. 
Adapting CLIP for medical VLP has shown significant promise. General medical models like PubMedCLIP \cite{eslami2023pubmedclip} focus on aligning representations with biomedical literature or refining the contrastive objective for radiology. 
In the ultrasound domain, specialized models \cite{MISHRA2025103611, xinjie_2025, li_2025, maani2025fetalclip, christensen2024vision, nature2025com} have demonstrated the clear benefit of domain-specific pre-training. 
While these approaches continually advance the field, existing frameworks, both general and specialized, predominantly adopt the standard binary contrastive objective. This formulation, however, presents two fundamental limitations when applied to ultrasound. First, the semantic ambiguity of Ultrasound reports, where diverse phrases describe similar findings, introduces noise for binary contrastive learners. Second, these models often lack the domain-specific structural priors to model the complex relationships between diagnostic attributes for clinical reasoning.
This motivates our work. Our Ultrasound-CLIP attempts to address these challenges, by introducing semantic soft labels and heterogeneous graph encoder to resolve semantic ambiguity and model structured clinical reasoning.

\vspace{-0.05in}

\section{US-365K Dataset}
\label{sec:dataset}
\vspace{-0.05in}

\begin{figure*}[th]
    \centering
    \includegraphics[width=\linewidth]{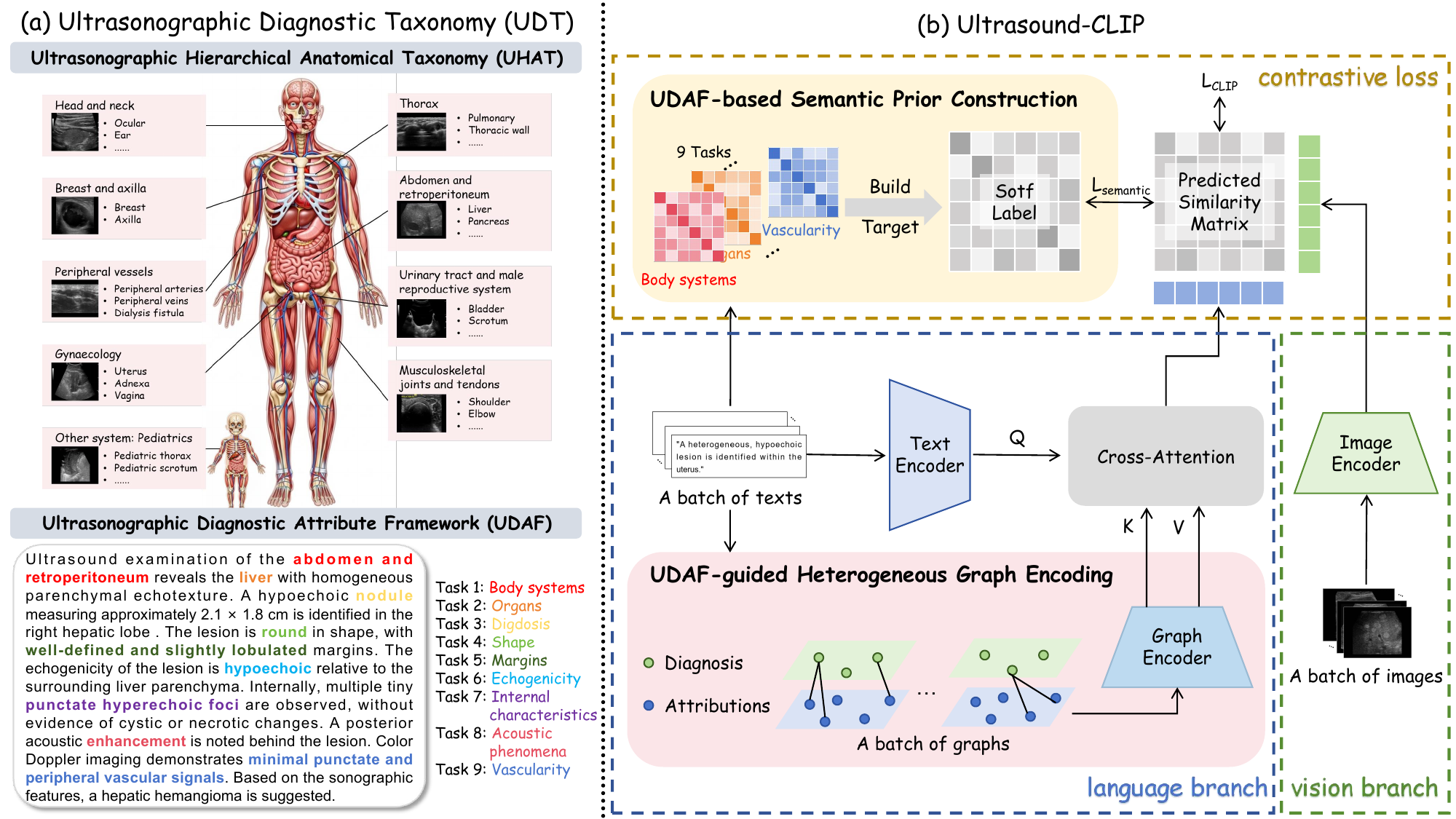}
    \caption{\textbf{Overview of UDT and Ultrasound-CLIP.} (a) The UDT serves as the semantic foundation, formalizing sonographic knowledge by standardizing anatomical hierarchies (UHAT) and defining 9 key diagnostic attributes (UDAF).
    (b) The Ultrasound-CLIP leverages UDT in two ways: (1) A UDAF-guided heterogeneous graph encoder fuses attribute relationships into the text embedding via cross-attention to model structured reasoning. (2) UDAF-based semantic priors are constructed to enable a dual-objective optimization that resolves ambiguity. The framework aligns visual features with these graph-enhanced, semantically-aware text representations.}
    \label{framework}
    \vspace{-7pt}
\end{figure*}

\begin{figure}[th]
\setlength{\abovecaptionskip}{0cm}
\setlength{\belowcaptionskip}{0cm}
    \centering
    \includegraphics[width=\linewidth]{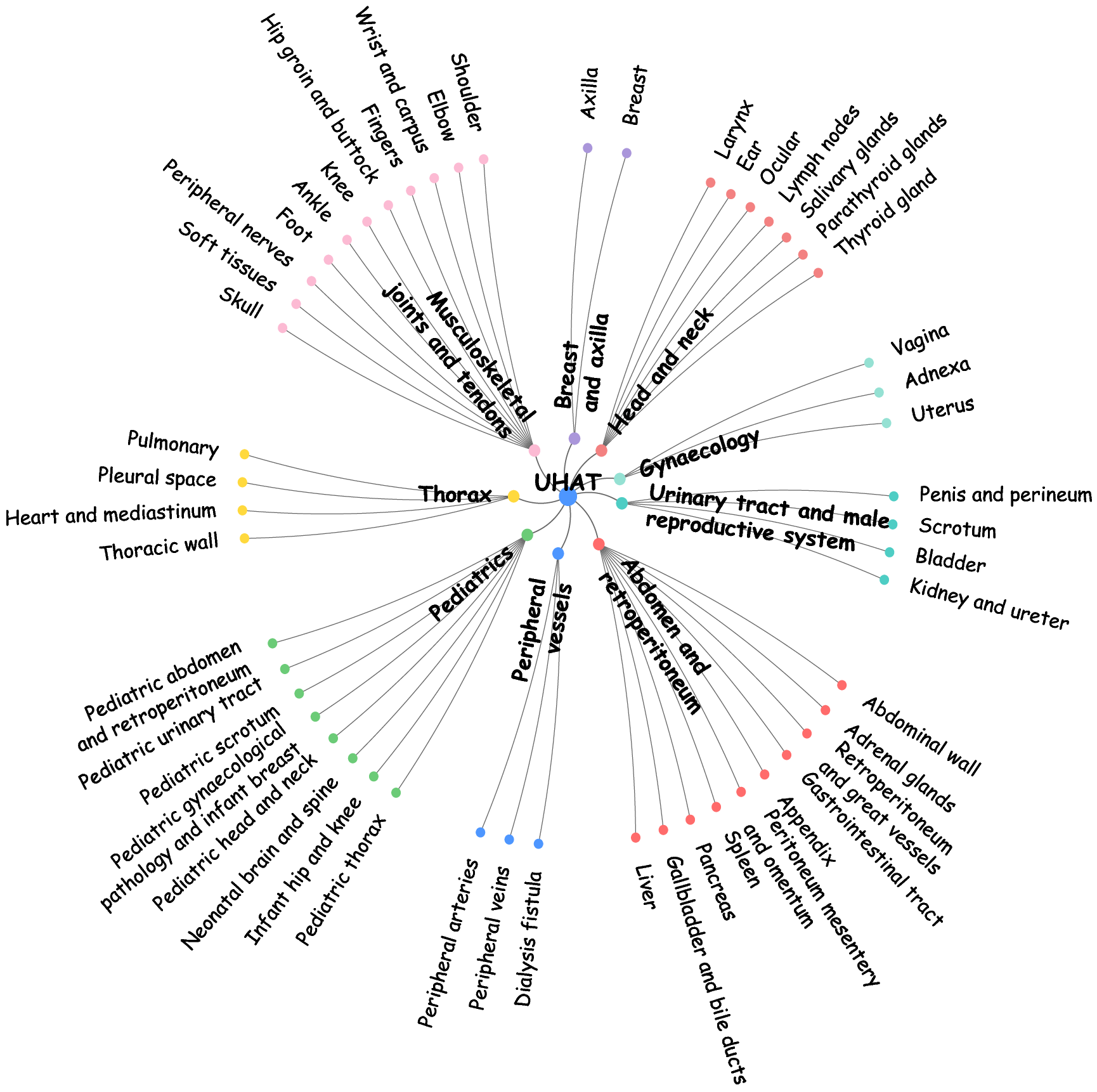}
    \caption{Visualization of US-365K's UHAT-based anatomical hierarchy.} 
    \label{datastatistics}
    \vspace{-0.25in}
\end{figure}

we develop the UDT, a unified knowledge framework that provides the semantic foundation for text normalization, knowledge-guided graph construction, and multimodal alignment (see Figure~\ref{framework}(a)). Based on UDT system, we construct the \textbf{US-365K} dataset, comprising 364,365 image–text pairs from 11,676 clinical cases.

\subsection{Ultrasonographic Diagnostic Taxonomy (UDT)}

The UDT is designed to standardize the complex anatomical and diagnostic semantics unique to ultrasound. It consists of two key components: 

\noindent\textbf{Ultrasonographic Hierarchical Anatomical Taxonomy (UHAT).} To resolve inconsistencies in anatomical categorization across diverse data sources, we introduce the UHAT, a unified two-level anatomical classification tree, organized by clinical principles into 9 \textit{body systems} and 52 \textit{organs}. Each organ node inherits contextual information from its parent system, thereby preserving the hierarchical dependency between local and systemic anatomy, and forming a robust foundation for standardized anatomical labeling. 
    
\noindent\textbf{Ultrasonographic Diagnostic Attribute Framework (UDAF).} While UHAT defines anatomical hierarchies, diagnostic interpretation in ultrasonography also relies on a structured set of lesion-related attributes. To capture this dimension, we establish the UDAF, a multi-dimensional diagnostic schema derived from standardized clinical reporting conventions. 
It formalizes nine diagnostic dimensions that clinicians commonly assess when interpreting ultrasound images, including \textit{body systems, organs, diagnosis, shape, margins, echogenicity, internal characteristics, acoustic phenomena and vascularity}. Each dimension is associated with a curated vocabulary of clinically valid descriptors (see Figure \ref{datastatistics}). This framework enables consistent text normalization and serves as the foundation for constructing heterogeneous lesion–attribute graphs.

By jointly leveraging UHAT and UDAF, the UDT offers a unified and comprehensive representation of ultrasonographic knowledge for mapping unstructured text to standardized, multi-dimensional labels.


\subsection{Data Collection, Construction and Annotation}
\textbf{Step 1: Multi-source Data Collection.} We collect ultrasound data from five publicly accessible repositories, ensuring diverse clinical coverage: \textit{Ultrasoundcaseinfo}\footnote{\url{https://www.ultrasoundcases.info/}} providing radiologist-curated teaching cases, \textit{LITFL 100+ Ultrasound Quiz}\footnote{\url{https://litfl.com/top-100/ultrasound/}} offering multiple diagnostic quiz, \textit{MedPix}\footnote{\url{https://medpix.nlm.nih.gov}} offering database hosted by the U.S. National Library of Medicine, \textit{POCUS Atlas}\footnote{\url{https://www.thepocusatlas.com/}} aggregating peer-reviewed point-of-care cases, and \textit{RadImageNet}\footnote{\url{https://app.radimagenet.com/}} providing open medical images. This multi-source approach ensures diversity in anatomy, pathology, and linguistic style. Full source details and licensing are provided in the Supplementary Material.

\noindent\textbf{Step 2: Data Construction.} We first filter all non-ultrasound content. To integrate the dynamic nature of sonography, video clips are decomposed into static frames at 0.5-second intervals, balancing temporal diversity and redundancy. 
%
For textual record, we extract structured label sets based on the UDAF taxonomy. The text-to-label extraction is implemented as a hybrid pipeline: structured prompts, designed under the UDAF schema, are issued to a large language model (Gemini~2.5~Flash) to automatically extract UDAF specific attributes. 

\noindent\textbf{Step 3: Image-text Pairing and Annotation.} Each refined image is systematically paired with its corresponding caption. For multi-image cases or composite figures, rule-based regular expressions matching are applied to identify subfigure indicators to automatically link subfigures with their respective subcaptions or narrative segments, ensuring fine-grained alignment between localized visual regions and descriptive text while preserving diagnostic context. Through this pipeline, we establish 364,365 image–text pairs from 11,676 clinical cases, including both static and video-derived ultrasound frames. The dataset captures extensive intra-organ variability and inter-system diversity, forming a multimodal corpus suitable for large-scale pretraining and downstream clinical analysis.

To ensure data quality and clinical reliability, all generated captions and diagnostic labels have been thorough reviewed by medical annotators and senior doctors. Ambiguous, inconsistent, or irrelevant cases are excluded. The dataset has an effective rate of over 90\%, with detailed procedures provided in the Supplementary Material.

\subsection{Data Statistics and Visualization}

We provide a comprehensive analysis of the US-365K dataset, which is characterized by its broad anatomical coverage, diverse pathological findings, and rich linguistic properties. 
As shown in Figure~\ref{datastatistics}, US-365K provides extensive coverage across all major anatomical regions defined by our UHAT. It captures a wide range of clinical contexts, spanning 9 high-level body systems and 52 fine-grained organ subcategories. 
Further analysis confirms the dataset's depth. 
The finding distribution ensures robust coverage of both common pathologies like ``fluid collection" and ``mass", as well as rarer diagnostic entities. The caption length distribution aligns with the brevity of medical reporting (typically $\leq$ 25 words), while a tail of longer captions provides linguistic diversity. Finally, the corpus vocabulary highlights a rich vocabulary that balances general and specialized medical language (e.g., ``cystic", ``hyperechoic"), making the dataset ideal for complex semantic understanding tasks. The full visualizations is provided in the Supplementary Material. 

\vspace{-0.05in}
\section{Ultrasound-CLIP}
\label{sec:model}
\vspace{-0.05in}


\subsection{Framework Overview}


As illustrated in Figure~\ref{framework}(b), our Ultrasound-CLIP framework is designed to overcome the challenges of semantic ambiguity and the lack of structural priors inherent in ultrasound data. It comprises a dual-encoder architecture ($f_\theta(\cdot)$ for images, $g_\phi(\cdot)$ for text) augmented with two key innovations: (1) a UDAF-guided Heterogeneous Graph Encoder ($h_\psi(\cdot)$) that models intra-text dependencies to create a semantically-aware text representation $\tilde{t}_i$, and (2) a UDAF-based Semantic Prior ($\tilde{\mathbf{S}}$) that provides fine-grained, non-binary supervision. These components are jointly optimized using a dual-objective that combines standard contrastive alignment with our novel semantic regularization loss.



\subsection{UDAF-based Semantic Prior Construction}
\label{sec:udaf}
To provide fine-grained domain knowledge beyond binary labels, we predefine $K$ UDAF task dimensions. For task $k$, we store a normalized label similarity matrix $\mathbf{S}^{(k)}\in\mathbb{R}^{L_k\times L_k}$, where $L_k$ is the label number in task $k$. 
Let $\mathcal{L}_i^{(k)}$ denote the label set of sample $i$ under task $k$. For any pair of samples $(i,j)$, we define the task-wise semantic affinity:
\begin{equation}
\label{eq:task_sim}
s_{ij}^{(k)} = \frac{1}{|\mathcal{L}_i^{(k)}|\,|\mathcal{L}_j^{(k)}|}
\sum_{a\in\mathcal{L}_i^{(k)}}\sum_{b\in\mathcal{L}_j^{(k)}}
\mathbf{S}^{(k)}_{ab}.
\end{equation}

We then aggregate task-wise affinities into an overall prior affinity:
\begin{equation}
\label{eq:prior}
\vspace{-0.1in}
\tilde{s}_{ij} \;=\; \frac{1}{K}\sum_{k=1}^{K} s_{ij}^{(k)}.
\end{equation}

Forming $\tilde{\mathbf{S}}=[\tilde{s}_{ij}]_{i,j=1}^B$ yields a $B\times B$ \emph{soft prior matrix}, where $B$ is the batch size. We enforce $\tilde{s}_{ii}=1$ for self-consistency. The prior $\tilde{\mathbf{S}}$ encodes continuous semantic similarity (soft labels) based on multi-task label agreement, and serves as a structured target to regularize the learned cross-modal similarities (see Sec.~\ref{sec:loss}).

\subsection{UDAF-guided Heterogeneous Graph Encoding}
\label{sec:graph}
To capture intra-text label dependencies, each textual annotation is converted into a sample-specific heterogeneous graph 
\(\mathcal{G}_i = (\mathcal{V}_i, \mathcal{E}_i)\), derived from the global UDAF label vocabulary \(\mathcal{V}^{\mathrm{UDAF}}\). 

\noindent\textbf{Graph Construction.} 
We define two disjoint node types based on the UDAF framework: diagnostic nodes \(\mathcal{V}^{(d)} \subset \mathcal{V}^{\mathrm{UDAF}}\) 
and attribute nodes \(\mathcal{V}^{(p)} = \mathcal{V}^{\mathrm{UDAF}}\setminus\mathcal{V}^{(d)}\). 
For a given sample \(i\), the active node subset \(\mathcal{V}_i \subseteq \mathcal{V}^{\mathrm{UDAF}}\) 
is determined by the label set \(\{\mathcal{L}_i^{(k)}\}_{k=1}^{K}\) extracted under the UDAF framework. 
Edges $\mathcal{E}_i$ are created as full bipartite connections between the sample's active diagnostic and attribute nodes: 
\vspace{-0.1in}
\begin{equation}
\mathcal{E}_i \;=\; \{(u,v)\mid u\in\mathcal{V}_i^{(d)},\ v\in\mathcal{V}_i^{(p)}\}.
\vspace{-0.1in}
\end{equation}





\noindent\textbf{Graph-text Fusion.} 
A lightweight heterogeneous GNN $h_\psi(\cdot)$ first computes node embeddings $\mathbf{Z}_i\in\mathbb{R}^{n_i\times D}$. We then apply an attention pooling to aggregate node embeddings into a fixed-length graph summary vector $\mathbf{g}_i \in\mathbb{R}^D$:
\begin{align}
\mathbf{a}_i &= \mathrm{softmax} \!\big(\mathbf{w}^\top \tanh(\mathbf{Z}_i W_a)\big) \in \mathbb{R}^{n_i},\nonumber\\
\label{eq:attnpool}
\mathbf{g}_i &= \mathbf{a}_i^\top \mathbf{Z}_i \in\mathbb{R}^D,
\end{align}
where $W_a$ and $\mathbf{w}$ are learnable. This pooled graph embedding $\mathbf{g}_i$, which encapsulates structured attribute relations, is then fused with the raw text embedding $t_i=g_\phi(t_i)$. We use a multi-head attention (MHA) where $t_i$ queries $\mathbf{g}_i$: 
\begin{equation}
\label{eq:cross_attn}
\vspace{-0.05in}
h_i = \mathrm{MHA}\big(Q=W_t t_i,\ K=W_g \mathbf{g}_i,\ V=W_g \mathbf{g}_i\big),
\end{equation}
where $W_t, W_g$ are projection matrices. Finally, we obtain the graph-enhanced text embedding $\tilde{t}_i\in\mathbb{R}^D$ via a gated residual connection: 
\begin{equation}
\label{eq:enhanced_text}
\vspace{-0.05in}
\tilde{t}_i = \mathrm{LayerNorm}\big(t_i + \alpha \tanh(h_i)\big), 0\le \alpha \le \alpha_{\max}.
\end{equation}

This stable fusion ensures that the final vector $\tilde{t}_i$ serves as the text-side representation for optimization.

\subsection{Dual-objective Optimization}
\label{sec:loss}

The overall objective of Ultrasound-CLIP is to learn aligned representations between ultrasound images and their graph-enhanced textual counterparts through a joint contrastive and semantic optimization strategy. 
Each image $x_i$ is encoded by a vision backbone $f_\theta(\cdot)$ into $x_i=f_\theta(x_i)\in\mathbb{R}^D$, while the corresponding text is encoded by $g_\phi(\cdot)$ and refined via the graph encoder to yield $\tilde{t}_i\in\mathbb{R}^D$. 
Both embeddings are L2-normalized before computing similarity. 
Given a training batch of size $B$, the predicted similarity matrix $\mathbf{P}\in\mathbb{R}^{B\times B}$ is defined as: 
\begin{equation}
\label{eq:pred_sim}
\vspace{-0.1in}
\mathbf{P}_{ij} \;=\; (x_i^\top \tilde{t}_j) / \tau,
\end{equation}
where $\tau>0$ is a learnable temperature parameter. 

\noindent\textbf{Contrastive Alignment Loss.}
We first employ a standard symmetric contrastive loss, $\mathcal{L}_{\mathrm{CLIP}}$, which enforces bidirectional image-to-text and text-to-image alignment by maximizing the agreement between positive pairs $(x_i, \tilde{t}_i)$ while discriminating against all other negative pairs in the batch. 
\begin{equation}
\label{eq:clip_loss}
\mathcal{L}_{\mathrm{CLIP}} = 
-\frac{1}{2B}\sum_{i=1}^{B}\Bigg(
\log \frac{e^{\mathbf{P}_{ii}}}{\sum_j e^{\mathbf{P}_{ij}}}
+
\log \frac{e^{\mathbf{P}_{ii}}}{\sum_j e^{\mathbf{P}_{ji}}}
\Bigg).
\end{equation}

\noindent\textbf{Semantic Alignment Loss.}
To address semantic ambiguity and leverage our structured UDAF priors, we introduce a novel semantic loss, $\mathcal{L}_{\mathrm{semantic}}$. This loss regularizes the predicted logits $\mathbf{P}$ to match our UDAF-based soft prior matrix $\tilde{\mathbf{S}}$. We compute the temperature-scaled distributions as $\mathbf{P}' = \mathrm{softmax}(\mathbf{P}/\tau)$ and $\mathbf{S}' = \mathrm{softmax}(\mathbf{S}^{\mathrm{prior}}/\tau)$. The semantic loss then combines an Mean Squared Error (MSE) term and a Kullback-Leibler (KL) divergence term: 
\begin{equation}
\label{eq:semantic_loss}
\vspace{-0.1in}
\begin{aligned}
\mathcal{L}_{\mathrm{semantic}}
= &\ \alpha_s \, \big\| \mathrm{clip}(\mathbf{P}) - \mathrm{norm}(\mathbf{S}^{\mathrm{prior}}) \big\|_F^2 \\
  & + (1-\alpha_s) \, D_{\mathrm{KL}}\big( \mathbf{P}' \;\big\|\; \mathbf{S}' \big),
\end{aligned}
\end{equation}
where $\alpha_s=0.6$ balances the terms, and $\mathrm{clip}(\cdot)$ and $\mathrm{norm}(\cdot)$ ensure numerical stability. The MSE term enforces a direct, value-to-value similarity match, while the $D{\mathrm{KL}}$ term enforces distributional consistency, ensuring that semantically related samples cluster appropriately.

\begin{table*}[th]
\centering
\caption{\textbf{Classification performance comparison across 9 ultrasound diagnostic attributes (T1–T9).} We report Accuracy (Acc), Recall, and their averages (AvgAcc, AvgRecall)). All values are reported in percentage (\%). * Models cannot be retrained due to unavailable code.}
\vspace{-0.1in}
\label{tab:classfication}
\resizebox{\textwidth}{!}{
\begin{tabular}{p{3.4cm}|
>{\centering\arraybackslash}p{1.2cm}>{\centering\arraybackslash}p{1.2cm}|
>{\centering\arraybackslash}p{1.2cm}>{\centering\arraybackslash}p{1.2cm}|
>{\centering\arraybackslash}p{1.2cm}>{\centering\arraybackslash}p{1.2cm}|
>{\centering\arraybackslash}p{1.2cm}>{\centering\arraybackslash}p{1.2cm}|
>{\centering\arraybackslash}p{1.2cm}>{\centering\arraybackslash}p{1.2cm}
}
\toprule
\textbf{Model} &
\multicolumn{2}{c|}{T1. Body System} &
\multicolumn{2}{c|}{T2. Organ} &
\multicolumn{2}{c|}{T3. Diagnosis} &
\multicolumn{2}{c|}{T4. Shape} &
\multicolumn{2}{c}{T5. Margins} \\
& Acc & Recall & Acc & Recall & Acc & Recall & Acc & Recall & Acc & Recall \\
\midrule
\multicolumn{11}{l}{\textit{General CLIP}} \\
\midrule
CLIP \cite{radford2021learning}  & 11.27 & 26.08 & 0.83 & 12.07 &4.89 & 29.22 & 6.26 & 12.74 & 28.91 & 52.28 \\
SigLIP \cite{zhai2023sigmoid} *& \textbf{35.34} & \textbf{46.91} & 7.90 & 28.16 & 33.25 & 38.95 & 24.46 & 24.93 & 42.94 & 51.47 \\
MetaCLIP \cite{ICLR2024_d1450d6c} & 5.99 & 29.14 & 2.58 & 14.83 & 4.48 & 24.54 & 3.67 & 13.37 & 29.63 & 58.50 \\
\midrule
\multicolumn{11}{l}{\textit{Medical CLIP}} \\
\midrule
MedCLIP \cite{wang2022medclip} *& 11.25 & 29.68 & 0.35 & 17.38 & 6.07 & 22.36 & 39.87 & 30.29 & 41.44 & 36.15 \\
PMC-CLIP \cite{lin2023pmc} & 16.70 & 34.71 & 9.37 & 20.97 & 24.26 & 33.32 & 8.94 & 22.78 & 61.88 & 57.92 \\
UniMed-CLIP \cite{khattak2024unimed} & 7.56 & 12.95 & 1.10 & 5.86 & 32.56 & 24.41 & 16.11 & 20.01 & 57.88 & 53.17 \\
BiomedCLIP \cite{zhang2023biomedclip} & 32.88 & 39.81 & 13.40 & 24.82 & 39.40 & 39.35 & 16.53 & 22.45 & 64.13 & 61.36 \\
\rowcolor{gray!20}
Ultrasound-CLIP-$D_{s+g}$ & 29.00 &	41.43 &	26.84 &	45.51 &	54.81 &	55.70 &	55.38 &	51.71 &	74.19 &	72.87  \\
\rowcolor{gray!20}
Ultrasound-CLIP-$D_{s}$ &  27.84 &	42.74 &	24.24 &	42.38 &	46.79 &	51.96 &	43.08 &	49.25 &	79.88 &	78.90 \\
\rowcolor{gray!20}
Ultrasound-CLIP-$D_{g}$ &  26.52 &	39.77 &	29.80 &	49.21 &	60.59 &	63.16 &	53.16 &	49.63 &	79.00 &	77.84 \\
\rowcolor{gray!20}
\textbf{Ultrasound-CLIP} & 32.27 & 44.11 & \textbf{31.49} & \textbf{51.27} & \textbf{64.05} & \textbf{66.15} & \textbf{70.04} & \textbf{64.11} & \textbf{84.44} & \textbf{84.42} \\
\bottomrule
\end{tabular}
}
\vspace{0.2cm}

\resizebox{\textwidth}{!}{
\begin{tabular}{p{3.4cm}|
>{\centering\arraybackslash}p{1.2cm}>{\centering\arraybackslash}p{1.2cm}|
>{\centering\arraybackslash}p{1.2cm}>{\centering\arraybackslash}p{1.2cm}|
>{\centering\arraybackslash}p{1.2cm}>{\centering\arraybackslash}p{1.2cm}|
>{\centering\arraybackslash}p{1.2cm}>{\centering\arraybackslash}p{1.2cm}|
>{\centering\arraybackslash}p{1.2cm}>{\centering\arraybackslash}p{1.2cm}
}
\toprule
\textbf{Model} &
\multicolumn{2}{c|}{T6. Echogenicity} &
\multicolumn{2}{c|}{T7. Internal Char.} &
\multicolumn{2}{c|}{T8. Posterior Acou.} & 
\multicolumn{2}{c|}{T9. Vascularity} &
\multicolumn{2}{c}{Average} \\
& Acc & Recall & Acc & Recall & Acc & Recall & Acc & Recall & AvgAcc & AvgRecall \\
\midrule
\multicolumn{11}{l}{\textit{General CLIP}} \\
\midrule
CLIP \cite{radford2021learning}  & 14.17 & 24.31 & 8.69 & 23.83 & 39.90 & 49.01 & 4.67 & 29.19 & 13.29 & 28.75 \\
SigLIP \cite{zhai2023sigmoid} *& 22.98 & 15.48 & 28.18 & 30.49 & 85.31 & 53.38 & 27.18 & 32.11 & 34.17 & 35.76 \\
MetaCLIP \cite{ICLR2024_d1450d6c} & 11.70 & 25.96 & 4.65 & 21.78 & 44.43 & 65.66 & 3.63 & 34.24 & 12.30 & 32.00 \\
\midrule
\multicolumn{11}{l}{\textit{Medical CLIP}} \\
\midrule
MedCLIP \cite{wang2022medclip} *& 30.55 & 30.42 & 16.61 & 49.18 & 79.71 & 47.01 & 2.44 & 24.47 & 25.37 & 31.88 \\
PMC-CLIP \cite{lin2023pmc} & 19.65 & 34.55 & 20.74 & 30.33 & 11.53 & 53.83 & 16.32 & 24.23 & 21.04 & 34.74 \\
UniMed-CLIP \cite{khattak2024unimed} & 15.84 & 21.21 & 0.01 & 43.93 & 34.45 & 43.18 & 27.45 & 24.78 & 21.44 & 27.72 \\
BiomedCLIP \cite{zhang2023biomedclip} & 22.15 & 26.57 & 19.67 & 26.27 & 85.41 & 54.43 & 10.73 & 20.97 & 33.81 & 35.11 \\
\rowcolor{gray!20}
Ultrasound-CLIP-$D_{s+g}$ &  51.29 & 51.03 & 42.27 &	50.53 &	63.64 &	59.45 &	60.19 &	47.57 &	50.84 &	52.87  \\
\rowcolor{gray!20}
Ultrasound-CLIP-$D_{s}$ &  48.81 &	54.04 &	41.68 &	48.05 &	74.31 &	63.24 &	51.00 &	47.48 &	48.62 &	53.12  \\
\rowcolor{gray!20}
Ultrasound-CLIP-$D_{g}$ &  56.50 &	58.37 &	45.17 &	54.32 &	51.44 &	58.16 &	46.65 &	45.62 &	49.87 &	55.12  \\
\rowcolor{gray!20}
\textbf{Ultrasound-CLIP} & \textbf{57.71} & \textbf{62.02} & \textbf{53.87} & \textbf{55.76} & \textbf{89.95} & \textbf{73.56} & \textbf{52.66} & \textbf{48.31} & \textbf{59.61} & \textbf{61.08} \\
\bottomrule
\end{tabular}
}
\vspace{-0.15in}
\end{table*}

The overall optimization objective combines both contrastive and semantic terms:
\begin{equation}
\label{eq:total_loss}
\vspace{-0.1in}
\mathcal{L} = \mathcal{L}_{\mathrm{CLIP}} + \lambda\, \mathcal{L}_{\mathrm{semantic}},
\end{equation}
where $\lambda$ balances the standard alignment objective with our domain-specific semantic regularization.
This dual-objective formulation allows the model to achieve domain-aware alignment while maintaining robust multimodal representations under weakly structured textual supervision.

\vspace{-0.1in}
\section{Experiment}
\label{sec:experiment}
\vspace{-0.05in}

We conduct extensive experiments to evaluate the performance of our proposed Ultrasound-CLIP. Our evaluation is designed to answer three key questions: (1) How does Ultrasound-CLIP perform on its native pre-training tasks (classification and retrieval) against state-of-the-art (SOTA) VLP models? (2) What is the specific contribution of our core components? (3) How well do the learned representations generalize to downstream diagnostic tasks under zero-shot (ZS), linear probe (LP), and fine-tuning (FT) settings?

\subsection{Datasets}
\vspace{-0.05in}
Our experiments utilize two dataset categories:
\begin{itemize}
    \item Pre-training Dataset: We use our large-scale US-365K (Sec.~\ref{sec:dataset}) for pre-training, in-domain classification, and retrieval evaluations.
    \item Downstream Datasets: To evaluate transferability, we use four public classification datasets: BUSBRA \cite{BUSBRA}, GIST514-DB \cite{he2023query2}, BreastMNIST \cite{yang2021medmnist}, and Breast \cite{ALDHABYANI2020104863}. These cover diverse ultrasound scenarios, including breast and gastrointestinal lesion discrimination.
\end{itemize}



\begin{table*}[th]
\centering
\footnotesize
\caption{\textbf{Retrieval performance (Recall@K) comparison on the US-365K test set.} We report both I2T and T2I results. * Models cannot be retrained due to unavailable code.}
\vspace{-0.1in}
\label{tab:retrieval}
\begin{tabular}{cc|ccc|ccc}
\toprule
\multicolumn{2}{c|}{\textbf{Method}} & \multicolumn{3}{c|}{I2T} & \multicolumn{3}{c}{T2I} \\
 & & R@5 & R@10 & R@50 & R@5 & R@10 & R@50 \\
\midrule
    ~& SigLIP \cite{zhai2023sigmoid} * & 0.0009 & 0.0016 & 0.0061 & 0.0011 & 0.0018 & 0.0079  \\
    General CLIP& CLIP \cite{radford2021learning}  & 0.142 & 0.2451 & 0.6306 & 0.1662 & 0.2783 & 0.6767  \\ 
    ~ & MetaCLIP \cite{ICLR2024_d1450d6c}  & 0.185 & 0.301 & 0.6922 & 0.1946 & 0.3148 & 0.7088  \\ 
\midrule
    ~ &MedCLIP \cite{wang2022medclip} * & 0.0001 & 0.0002 & 0.0007 & 0.0001 & 0.0002 & 0.001  \\
    Medical CLIP  & UniMed-CLIP \cite{khattak2024unimed}& 0.1566 & 0.2667 & 0.6734 & 0.1551 & 0.2641 & 0.6798  \\ 
    ~ & BiomedCLIP \cite{zhang2023biomedclip}  & 0.1788 & 0.2979 & 0.7029 & 0.1864 & 0.3089 & 0.7206  \\ 
    ~ & PMC-CLIP \cite{lin2023pmc} & 0.1808 & 0.3011 & 0.7215 & 0.1814  & 0.3038  & 0.7312  \\ 
\midrule
    \rowcolor{gray!20}
    ~ & Ultrasound-CLIP-$D_{s+g}$ & 0.1493 & 0.2542 & 0.6449 & 0.1518 & 0.2606 & 0.6683  \\ 
    \rowcolor{gray!20}
    Ultrasound-CLIP & Ultrasound-CLIP-$D_{s}$ & 0.1568 & 0.2683 & 0.6692 & 0.155 & 0.2659 & 0.6707  \\ 
    \rowcolor{gray!20}
    ~ & Ultrasound-CLIP-$D_{g}$ & 0.2147 & 0.3444 & 0.7638 & 0.2147 & 0.352 & 0.7774  \\ 
    \rowcolor{gray!20}
    ~ & \textbf{Ultrasound-CLIP} & \textbf{0.2359} & \textbf{0.3745} & \textbf{0.7909} & \textbf{0.2383} & \textbf{0.3781} & \textbf{0.8022} \\ 
\bottomrule
\end{tabular}
\vspace{-0.1in}
\end{table*}

\begin{table*}[ht]
\centering
\caption{\textbf{Downstream task generalization comparison across 4 ultrasound datasets.} We report the performance under ZS, LP, and FT settings. All values are reported in percentage (\%). * Models cannot be retrained due to unavailable code.}
\vspace{-0.1in}
\label{tab:recall_comparison}
\footnotesize
\begin{tabular}{l|ccc|ccc|ccc|ccc}
\toprule
\textbf{Method} & \multicolumn{3}{c|}{\textbf{BUS-BRA}} & \multicolumn{3}{c|}{\textbf{GIST514-DB}} & \multicolumn{3}{c|}{\textbf{BreastMNIST}} & \multicolumn{3}{c}{\textbf{Breast}} \\
 & ZS & LP & FT & ZS & LP & FT & ZS & LP & FT & ZS & LP & FT \\
\midrule
\multicolumn{11}{l}{\textit{General CLIP}} \\
\midrule
CLIP \cite{radford2021learning} & 49.82 & 54.56 & 82.16 & 50.18 & 62.02 & 76.06 & 39.84 & 51.19 & 85.16 & 63.42 & 54.74 & 83.54 \\
SigLIP \cite{zhai2023sigmoid} * & 50.47 & 55.59 & \textbf{83.69} & 47.51 & 66.67 & \textbf{81.52} & \textbf{54.30} & 60.71 & 87.97 & 41.52 & 64.67 & 85.09 \\
MetaCLIP \cite{ICLR2024_d1450d6c} & 52.73 & 50.42 & 82.50 & 48.26 & 57.41 & 73.25 & 46.27 & 51.19 & 86.78 & 49.66 & 51.76 & 82.76 \\
\midrule
\multicolumn{11}{l}{\textit{Medical CLIP}} \\
\midrule
PMC-CLIP \cite{lin2023pmc}& 49.04 & 65.40 & 79.57 & 49.17 & 63.43 & 68.41 & 50.51 & 61.90 & 81.89 & 51.22 & 62.76 & 79.81 \\
MedCLIP \cite{wang2022medclip} * & 49.23 & 57.60 & 82.52 & 50.22 & 68.19 & 71.96 & 39.28 & 60.71 & 78.76 & 64.95 & 50.25 & 86.29 \\
UniMed-CLIP \cite{khattak2024unimed} & 53.68 & 71.14 & 82.16 & 47.78 & \textbf{69.48} & 74.52 & 46.85 & 78.01 & \textbf{90.35} & \textbf{71.70} & 77.01 & 90.28 \\
BiomedCLIP \cite{zhang2023biomedclip} & 50.41 & 71.71 & 79.57 & 50.42 & 68.82 & 74.59 & 35.79 & 78.95 & 85.79 & 64.90 & 77.08 & 81.87 \\
\textbf{Ultrasound-CLIP} & \textbf{57.02} & \textbf{72.47} & 82.99 & \textbf{54.18} & 68.19 & 72.62 & 39.16 & \textbf{82.33} & 89.16 & 67.32 & \textbf{78.62} & \textbf{92.13} \\
\bottomrule
\end{tabular}
\vspace{-0.15in}
\end{table*}

\vspace{-0.05in}
\subsection{Implementation Details}
\vspace{-0.05in}
Our framework is implemented in PyTorch and trained on 4 NVIDIA A100 GPUs. The image encoder is a ViT \cite{dosovitskiy2021an} architecture, the text encoder is BioClinical-BERT \cite{lee2020biobert}, and the GNN is heterogeneous graph transformer\cite{hu2020heterogeneous}. 
The US-365K dataset is divided into training, validation, and test datasets in a 6:2:2 ratio.
All models are trained for 50 epochs using the AdamW optimizer with $\beta_1 = 0.9$, $\beta_2 = 0.999$, a batch size of 128, and a learning rate of $5 \times 10^{-4}$. The temperature parameters $\tau_1$, $\tau_2$, and $\tau_3$ are set to 0.07, and the flexibility weight $\lambda$ is set to 0.2. The parameter $\alpha_{\text{max}}$ is set to 0.2. 
Full details are provided in the Supplementary Material. 


\vspace{-0.05in}
\subsection{Evaluation on US-365K (Pre-training Tasks)}
\vspace{-0.05in}
\noindent\textbf{Multi-task Ultrasound Classification.} As shown in Table~\ref{tab:classfication}, Ultrasound-CLIP establishes a new state-of-the-art by a significant margin. General CLIP models perform poorly, confirming the existence of a profound modality gap. While medical CLIP models show moderate improvements, they remain inconsistent. In contrast, our Ultrasound-CLIP achieves an average accuracy of 59.61\%, outperforming the best baseline (BiomedCLIP, 33.81\%) by over 25 percentage points. Notably, our model excels in fine-grained and critical attributes like \textit{Diagnosis} (64.05\% vs. 39.40\%) and \textit{Margins} (84.44\% vs. 64.13\%). This demonstrates that our semantic loss and graph encoding successfully capture the specialized, fine-grained semantics of ultrasound diagnostics that other models miss.


\noindent\textbf{Image-text Retrieval.} We evaluate image-to-text (I2T) and text-to-image (T2I) retrieval performance on the US-365K test set (Table~\ref{tab:retrieval}). General and medical CLIP models show modest performance, with the best baseline (PMC-CLIP) achieving 0.3011 at R@10 (I2T). Our full Ultrasound-CLIP model significantly outperforms all baselines, achieving 0.3745 at R@10 (I2T) and 0.8022 at R@50 (T2I). This confirms that our model learns a more robustly aligned cross-modal embedding space.

\noindent\textbf{Ablation Studies on Core Components.}
To validate the efficacy of our proposed components, we conduct rigorous ablation studies on both classification (Table~\ref{tab:classfication}) and retrieval tasks (Table~\ref{tab:retrieval}). Specifically, we examine the contributions of the UDAF-guided semantic loss and the heterogeneous graph encoder through controlled variants. 
Removing both components (\textit{Ultrasound-CLIP-$D_{s+g}$}) results in performance comparable to standard CLIP (e.g., I2T R@10: 0.2542), indicating that simply training on US-365K is insufficient. 
Removing the UDAF-guided semantic loss (\textit{Ultrasound-CLIP-$D_s$}) yields a marginal improvement (e.g., I2T R@10: 0.2683), suggesting that the graph encoder alone cannot fully resolve complex cross-modal alignments. 
In contrast, removing the heterogeneous graph encoder (\textit{Ultrasound-CLIP-$D_g$}) leads to a more noticeable improvement (e.g., I2T R@10: 0.3444), highlighting the strong guidance provided by semantic supervision. 
Consistently, similar trends are observed in classification, where the full model achieves the best AvgAcc/AvgRecall, demonstrating that both components contribute complementary benefits. 
The complete model achieves the best performance (e.g., I2T R@10: 0.3745), confirming that the graph encoder enhances structured representation learning, while the UDAF-guided semantic loss provides fine-grained cross-modal alignment.

\begin{figure}[t]
\centering
\subfigure[without graph encoder]{ 
\begin{minipage}{0.4\textwidth}
\centering
\includegraphics[width = 0.75\textwidth]{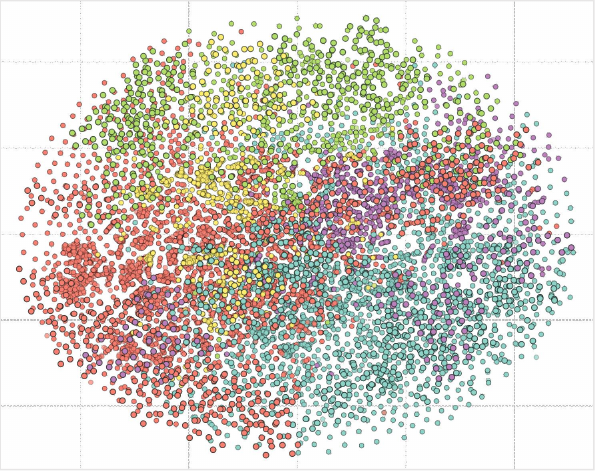} 
\end{minipage}}
\vspace{-0.2in}
\subfigure[with graph encoder]{ 
\begin{minipage}{0.4\textwidth}
\centering
\includegraphics[width = 0.75\textwidth]{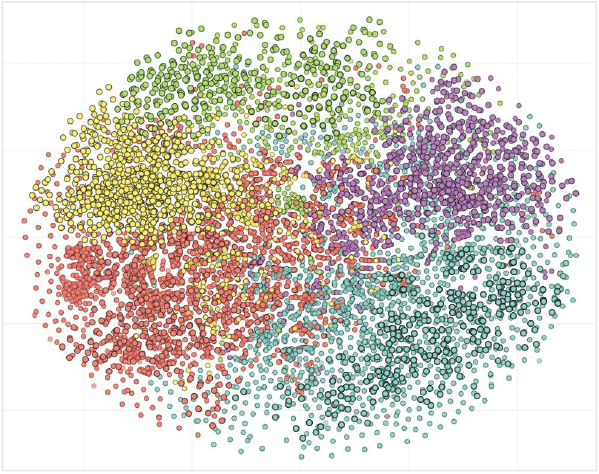} 
\end{minipage}}
\caption{t-SNE visualization of text embeddings without and with our UDAF-guided graph encoder, with categories colored by diagnosis: \textit{fluid collection} (blue), \textit{mass} (purple), \textit{normal appearance} (red), \textit{cyst} (green), and \textit{nodule} (yellow). }
\label{tsne}
\vspace{-0.2in}
\end{figure}

\vspace{-0.05in}
\subsection{Generalization to Downstream Tasks}
\vspace{-0.05in}
Finally, we assess the transferability of the learned representations by evaluating on four downstream classification tasks under ZS, LP, and FT settings. 
Table~\ref{tab:recall_comparison} summarizes the results. Our Ultrasound-CLIP demonstrates superior generalization across all settings. In the Zero-Shot setting, our model achieves an average performance of 54.42\%, drastically outperforming models like UniMed-CLIP (49.44\%) and BiomedCLIP (50.38\%). This highlights that our pre-training on US-365K imbues the model with robust, generalizable diagnostic knowledge. In the Linear Probe setting, our model (75.40\%) again surpasses all baselines, including the highly performing BiomedCLIP (74.14\%). This confirms our features are more linearly separable and semantically rich. In the Full Fine-Tuning setting, Ultrasound-CLIP (84.23\%) remains highly competitive, achieving the best or second-best performance on all datasets, notably peaking at 92.13\% on the Breast dataset.

\begin{figure}[t]
\setlength{\abovecaptionskip}{0cm}
    \centering
    \includegraphics[width=\linewidth]{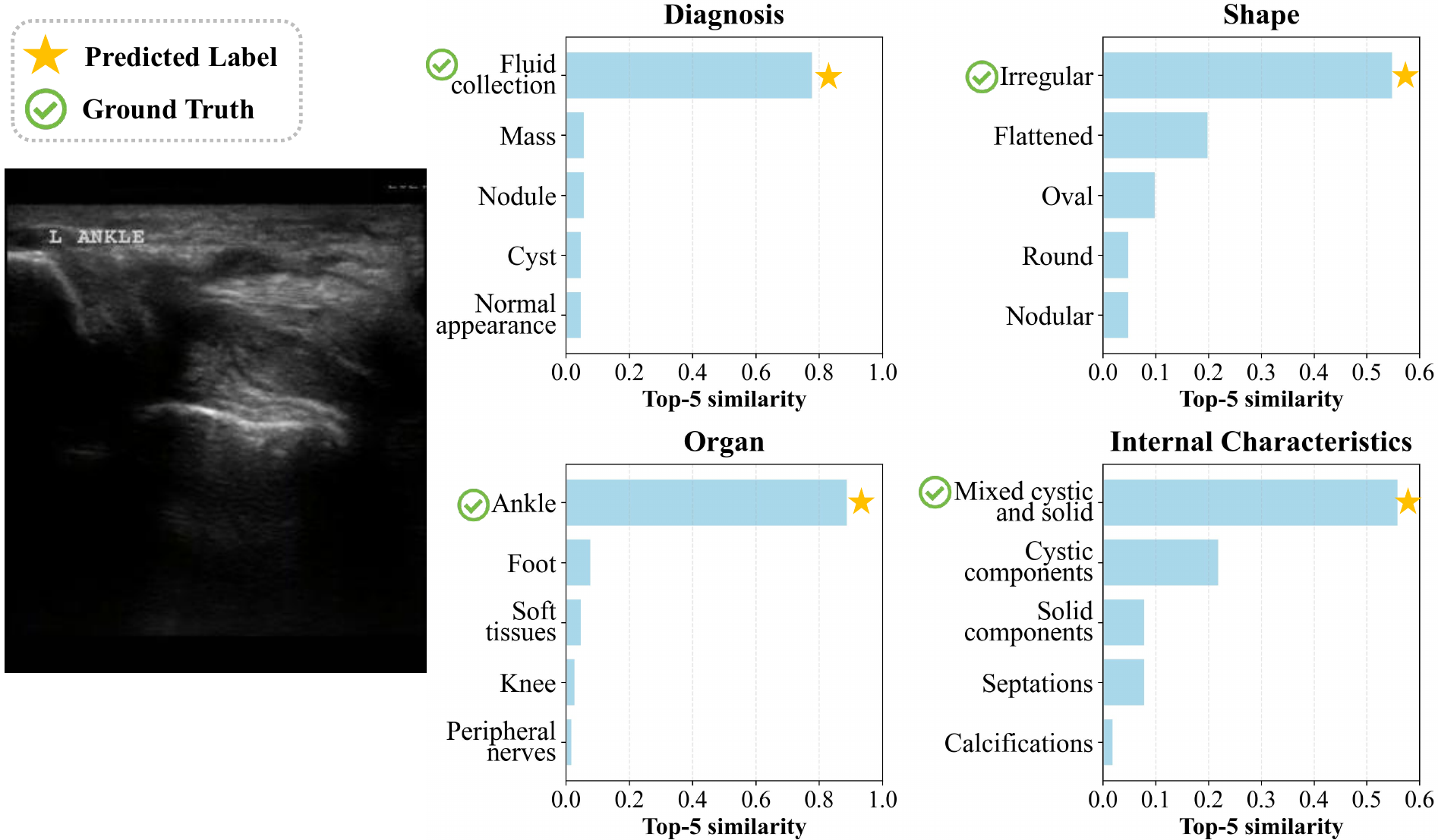}
    \caption{Visualization of diagnostic reasoning for left ankle ultrasound showing strong clinical coherence.}
    \label{case}
    \vspace{-0.2in}
\end{figure}

\vspace{-0.05in}
\subsection{Qualitative Analysis and Visualization}
\vspace{-0.05in}
\noindent\textbf{Graph Encoder Enhances Semantic Disambiguation.} To verify that our graph encoder improves representation quality, we visualize the t-SNE projections of text embeddings. Embeddings from the baseline text encoder (see Figure~\ref{tsne}(a)) show significant cluster overlap. In contrast, after fusion with our UDAF-guided graph representations (see Figure~\ref{tsne}(b)), the clusters become substantially more distinct and semantically coherent. This qualitatively demonstrates that our graph encoder successfully disambiguates fine-grained clinical concepts, leading to a more discriminative representation space.


\noindent\textbf{Case Study: Diagnostic Reasoning.} We further illustrate the interpretability of Ultrasound-CLIP through a case study in Figure~\ref{case}. The model demonstrates strong clinical coherence, correctly identifying the primary diagnosis as ``Fluid collection". More importantly, its top-ranked predictions for supporting attributes are all clinically consistent with this diagnosis, such as ``Hypoechoic" for echogenicity. This case-level analysis confirms that our UDAF-guided framework moves beyond simple keyword matching and learns the complex, structured relationships between diagnostic attributes, mirroring clinical reasoning.
\vspace{-0.05in}

\section{Conclusion}
\label{sec:conclusion}
\vspace{-0.05in}
In this paper, we address the fundamental challenges in ultrasound VLP: a profound dataset gap, pervasive semantic ambiguity, and a lack of structured clinical reasoning. 
Our work provides a unified data and model solution. We first introduce introduced the UDT, a novel knowledge framework used to construct US-365K, the first large-scale, pan-anatomy ultrasound benchmark. We then proposed Ultrasound-CLIP, a modality-tailored framework that integrates UDAF-guided semantic soft labels and a heterogeneous graph encoder to resolve semantic ambiguity and model clinical reasoning. 
Extensive experiments validate our approach. Our framework sets a new state-of-the-art, significantly outperforming all baselines on classification, retrieval, and downstream generalization tasks. By providing the field's first large-scale benchmark and a novel semantic-aware model, this work paves the way for a new class of powerful, clinically-aware foundation models in ultrasonography.
\section{Acknowledge}
\label{sec:acknowledge}

This work was supported by the National Key R\&D Program of China (No.2025YFG0100700), the ``Pioneer" and ``Leading Goose" R\&D Program of Zhejiang (No.2025C02068), the Hong Kong Research Grants Council (RGC) Early Career Scheme grant (No.22203525), the Natural Science Foundation of Zhejiang Province (No.LTGG24F020002, LY24F020013), the National Natural Science Foundation of China (No. 62576304, 62102349), and the Scientific Research Cultivation Fund of Hangzhou City University (No.J-202404).
The authors would like to acknowledge the Supercomputing Center of Hangzhou City University, for the support of advanced computing resources.

{
    \small
    \bibliographystyle{ieeenat_fullname}
    \bibliography{main}
}

\clearpage
\setcounter{page}{1}
\maketitlesupplementary

\renewcommand{\thesection}{\Alph{section}} 
\renewcommand{\thesubsection}{\thesection.\arabic{subsection}} 

\setcounter{section}{0} 

\section{Ultrasonographic Diagnostic Attribute Framework (UDAF)}
\label{UDTUDAF}
\begin{figure}[th]
    \centering
    \includegraphics[width=\linewidth]{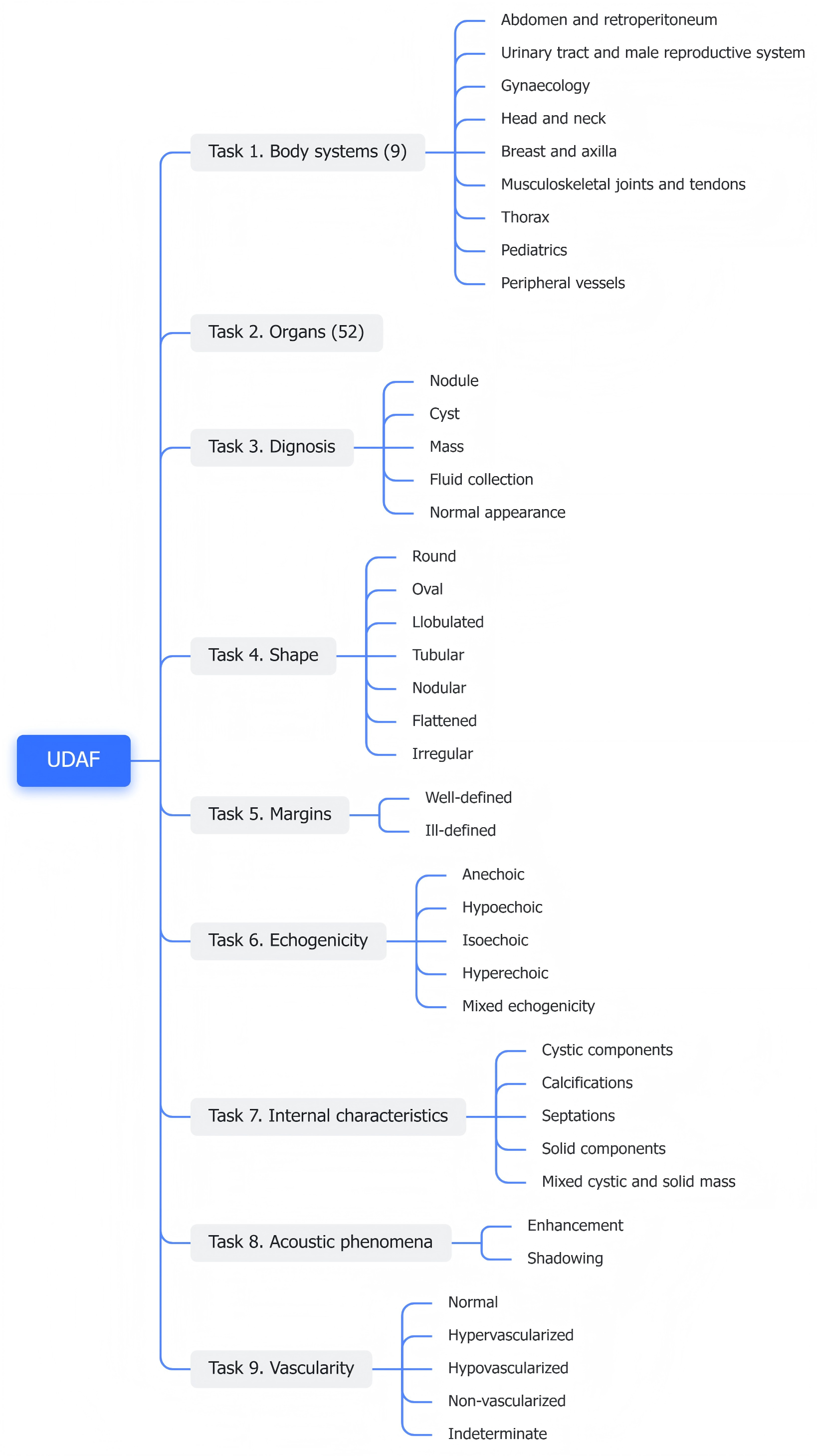}
    \caption{\textbf{Overview of the UDAF schema.} The framework decomposes ultrasound interpretation into nine standardized diagnostic dimensions. Note that Task 2 details are visualized in Figure~\ref{datastatistics} of the main paper.}
    \label{UDAF}
\end{figure}
The Ultrasonographic Diagnostic Attribute Framework (UDAF) serves as the semantic backbone of our dataset. As illustrated in Figure~\ref{UDAF}, UDAF provides a unified taxonomy that standardizes anatomical, morphological, and pathological attributes across heterogeneous ultrasound systems. By decomposing diagnostic-relevant information into hierarchical, multi-level attributes, UDAF enables consistent and granular annotation across diverse clinical cases.

\section{Data Processing Pipeline}
\label{dataprocess}
\begin{figure*}[th]
    \centering
    \includegraphics[width=\linewidth]{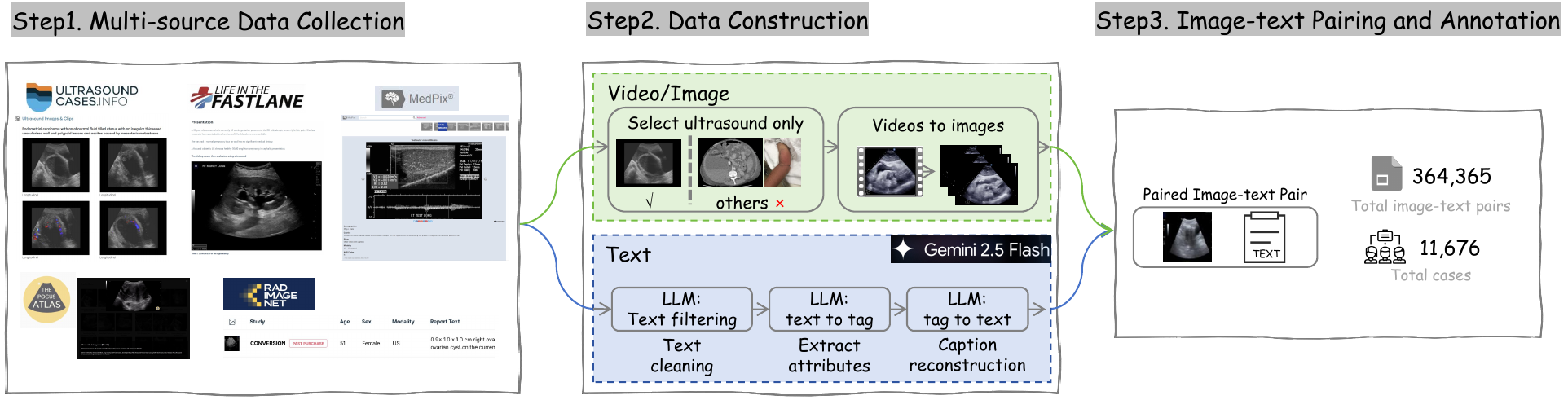}
    \caption{\textbf{The US-365K data construction pipeline.}}
    \label{appedixdataprocess}
\end{figure*}

To ensure the reproducibility, transparency, and completeness of dataset construction, we provide a detailed description of our data construction pipeline, which consists of three sequential stages: (1) multi-source data collection, (2) data construction, and (3) image–text pairing and annotation.
The complete workflow is illustrated in Figure ~\ref{appedixdataprocess}, which highlights how heterogeneous online resources are transformed into a unified, clinically reliable ultrasound corpus. 

\textbf{Step 1: Multi-source Data Collection.} To ensure comprehensive coverage and clinical diversity of ultrasound data, we systematically collect cases from five publicly accessible repositories, each providing distinct yet complementary characteristics and content types. All data collection adheres strictly to the terms of service and licensing agreements of the respective platforms, with all sources explicitly permitting academic and research use. The data sources are summarized as follows. 
\begin{itemize}
    \item \textit{UltrasoundCases}\footnote{\url{https://www.ultrasoundcases.info/}}. This continuously maintained teaching-oriented platform developed by radiologists and sonographers provides a wide range of ultrasound imaging cases categorized by anatomical systems and organs. The website organizes data into 10 body systems and 60 organs, which served as the primary reference for constructing our HATU. The site exclusively contains ultrasound cases with rich visual and textual descriptions. In total, we collect 50,950 ultrasound images paired with corresponding textual annotations.
    \item \textit{LITFL 100+ Ultrasound Quiz}\footnote{\url{https://litfl.com/top-100/ultrasound/}}. The Life in the Fast Lane ultrasound quiz collection presents self-assessment materials comprising clinical scenarios, diagnostic questions, images, and key learning points. Although the number of ultrasound examples is limited, the accompanying text provides rich diagnostic reasoning and context. We select 111 representative cases from this source, including 213 videos and 102 still images, to enhance linguistic diversity and narrative quality.
    \item \textit{MedPix}\footnote{\url{https://medpix.nlm.nih.gov}}. The MedPix database, hosted by the U.S. National Library of Medicine, offers a large-scale, open-access collection organized by disease, anatomy, pathology, and imaging modality. We retrieve 40 ultrasound cases with 109 associated images and corresponding textual descriptions. These cases feature well-structured diagnostic summaries and cross-references, contributing to the medical accuracy of our corpus.
    \item \textit{POCUS Atlas}\footnote{\url{https://www.thepocusatlas.com/}}. This collaborative educational platform aggregates peer-reviewed point-of-care ultrasound (POCUS) cases submitted by clinicians worldwide. The website covers 13 organ categories and provides annotated imaging content with brief clinical notes. Due to limited textual descriptions, we primarily utilize 1,190 GIF sequences to enrich the dataset with dynamic ultrasound examples.
    \item \textit{RadImageNet}\footnote{\url{https://app.radimagenet.com/}}. RadImageNet is a GitHub-hosted open medical imaging resource. We incorporated the subset of publicly available ultrasound images for academic use. Although textual descriptions are sparse, the dataset provides standardized, high-quality images that help balance underrepresented organ classes. We ultimately include 3,000 image-text pairs from this source. All collected data are licensed under Creative Commons or equivalent open-access frameworks that explicitly permit academic research and educational use, enabling reproducible research within the scientific community.
\end{itemize}

\textbf{Step 2: Data Construction.} We first filter all non-ultrasound content to ensure strict domain relevance, retaining only ultrasound-specific videos and images. To integrate dynamic examinations, video clips are decomposed into static frames at 0.5-second intervals, balancing temporal diversity and redundancy. Although video samples account for a relatively small portion of the data, the extracted frames enrich the dataset with temporal cues such as probe motion, tissue deformation, and transient lesion appearances—capturing the dynamic nature of real-world sonographic examinations.

For textual refinement, we adopt a hybrid strategy combining large language model (LLM)–based automation and expert supervision. Specifically, we design structured prompts for Gemini 2.5 Flash to guide multi-stage text cleaning and reconstruction. The process begins with automatic filtering to remove irrelevant, redundant, or noisy descriptions, followed by medical attribute extraction to obtain structured entities such as anatomy, modality, and diagnostic findings. Subsequently, the extracted tags are reformulated into coherent and concise diagnostic captions through a “tag-to-text” generation step. All outputs are manually reviewed by trained annotators to ensure clinical accuracy, readability, and alignment with standard ultrasound reporting conventions.

\textbf{Step 3: Image-text Pairing and Annotation.} Following the standardized image and text refinement, each processed ultrasound image is explicitly aligned with its corresponding caption through a systematic pairing procedure. For multi-image cases or composite figures, we apply rule-based regular expression matching to detect subfigure indicators (e.g., ``(A)”, ``(a)”) and automatically associate each subfigure with its respective subcaption or narrative segment. This ensures fine-grained correspondence between localized visual regions and descriptive text, preserving diagnostic context across subcomponents.

Through this alignment pipeline, we establish 364,365 image–text pairs from 11,676 clinical cases, encompassing both static and video-derived ultrasound frames. The resulting dataset captures rich intra-organ variability and inter-system diversity, forming a coherent multimodal corpus suitable for large-scale pretraining and downstream clinical analysis.

\section{Data Quality}
\label{dataquality}

\begin{table}[t]
\centering
\caption{\textbf{Expert confidence ratings.} Self-reported confidence scores (1--5) from the three verifying medical experts across anatomical specialties. Higher values indicate stronger domain expertise.}
\label{tab:specialty_confidence}
\begin{tabular}{lccc}
\toprule
\textbf{Specialty} & \textbf{Expert 1} & \textbf{Expert 2} & \textbf{Expert 3} \\
\midrule
Liver                               & 4 & 5 & 4 \\
Kidney                              & 4 & 4 & 5 \\
Bladder                             & 4 & 3 & 4 \\
Scrotum                             & 3 & 4 & 4 \\
Uterus                              & 4 & 4 & 3 \\
Adnexa                              & 3 & 4 & 4 \\
Thyroid                             & 5 & 4 & 4 \\
Lymph nodes                         & 4 & 4 & 4 \\
Breast                              & 4 & 3 & 4 \\
Shoulder                            & 3 & 4 & 3 \\
Knee                                & 3 & 4 & 4 \\
Pleural space                       & 4 & 4 & 4 \\
Lung                                & 4 & 4 & 4 \\
Neonatal brain                      & 3 & 3 & 4 \\
\bottomrule
\end{tabular}
\end{table}

To ensure the reliability and consistency of annotations across different anatomical systems, we implement a structured multi-stage quality control procedure aligned with the dataset construction workflow. During the annotation phase, three medical annotators examine outputs from the data curation pipeline, including the filtering of non-ultrasound content, the extraction of video frames at 0.5-second intervals, and the automatic UDAF-based label generation. Annotators verify the accuracy of extracted labels, the completeness of case information, and the correctness of rule-based subfigure matching used in image–text pairing. Cases with ambiguous anatomy, inconsistent textual records, or uncertain visual findings are flagged for further review.

In the review phase, three medical experts independently evaluate the refined image–text pairs. Each reviewer receives the ultrasound image, its caption, and the UDAF-aligned structured label set, and assesses two dimensions: \textit{semantic alignment} (whether the caption accurately describes the acquired visual information) and \textit{diagnostic consistency} (whether the labels match the clinical semantics expressed in the caption and image). To ensure broad anatomical coverage, samples from all nine ultrasound systems are included in the review pool. Furthermore, a dedicated confidence assessment is conducted in which experts rate their familiarity and interpretive confidence across 14 representative subspecialties, selected to provide balanced coverage while maintaining conciseness. The confidence scores, ranging from 1 to 5, are summarized in Table \ref{tab:specialty_confidence}.

Pairs for which two or more experts identify issues in alignment or consistency are sent to a consensus adjudication stage. Annotators and experts collaboratively refine captions or labels until agreement is achieved. Across a randomly sampled subset of 5,000 image–text pairs from the dataset, the effective quality rate exceeds 93.2\%.

\section{US-365K Dataset Statistics}
\label{datasta}



\begin{figure}[th]
    \centering
    \subfigure[Distribution of diagnostic findings.]{ 
    \begin{minipage}{0.5\textwidth}
    \centering
    \includegraphics[width = 1\textwidth]{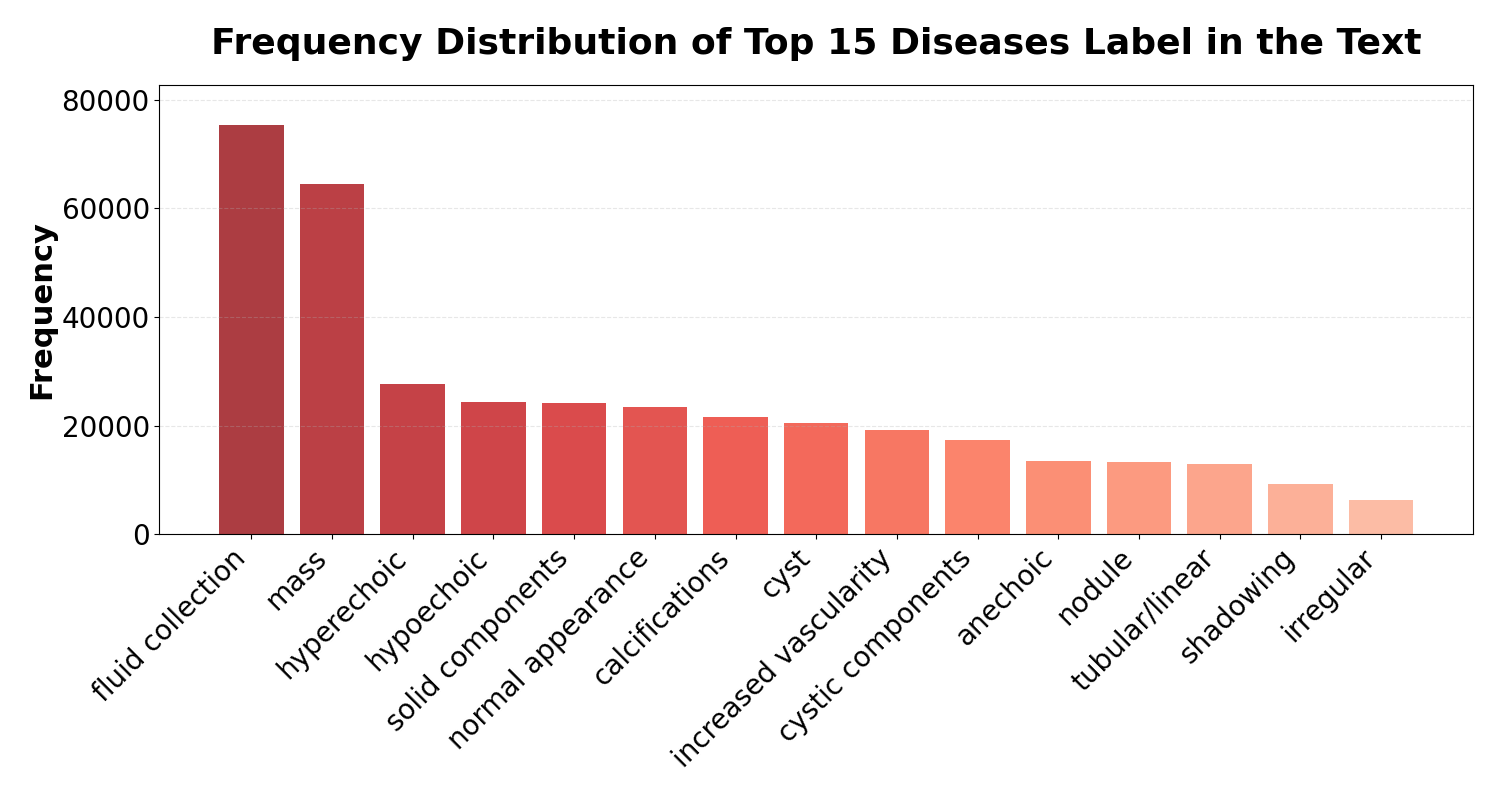} 
    \end{minipage}}
    
    \subfigure[Word cloud of captions.]{ 
    \begin{minipage}{0.5\textwidth}
    \centering
    \includegraphics[width = 1\textwidth]{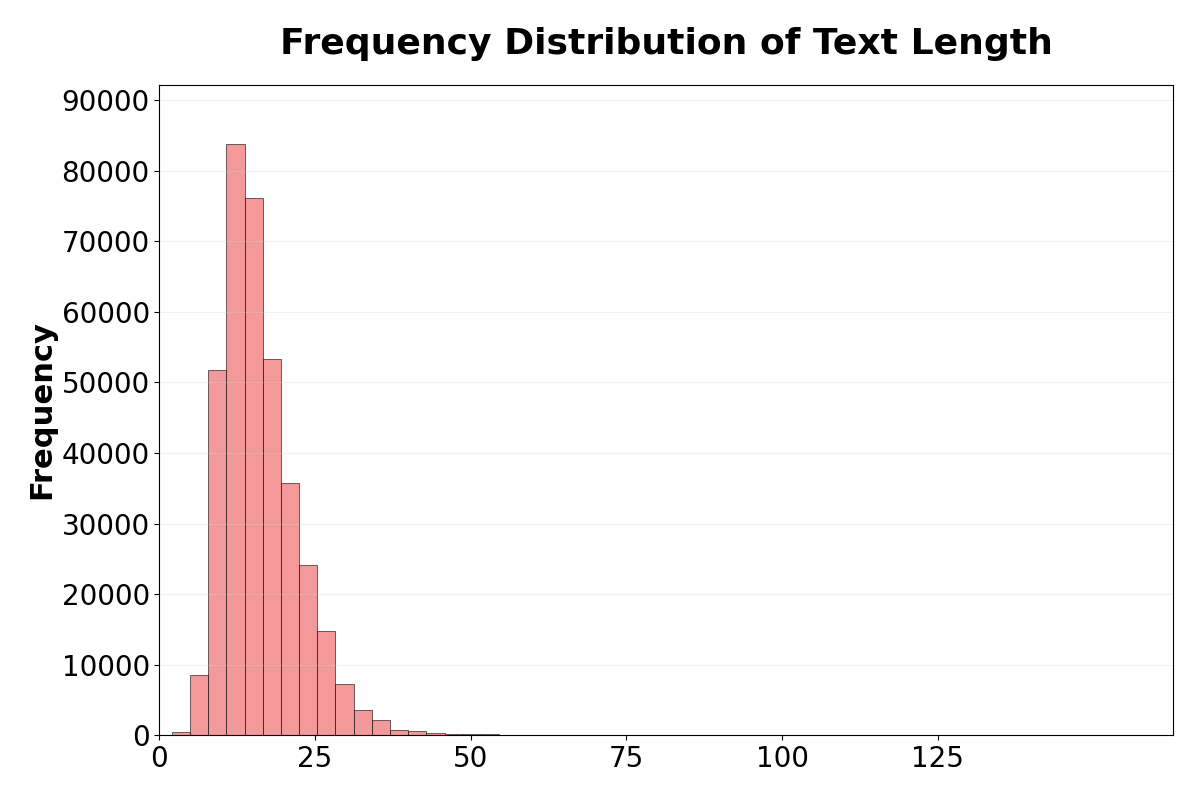} 
    \end{minipage}}

    \subfigure[Caption length distribution.]{ 
    \begin{minipage}{0.5\textwidth}
    \centering
    \includegraphics[width = 1\textwidth]{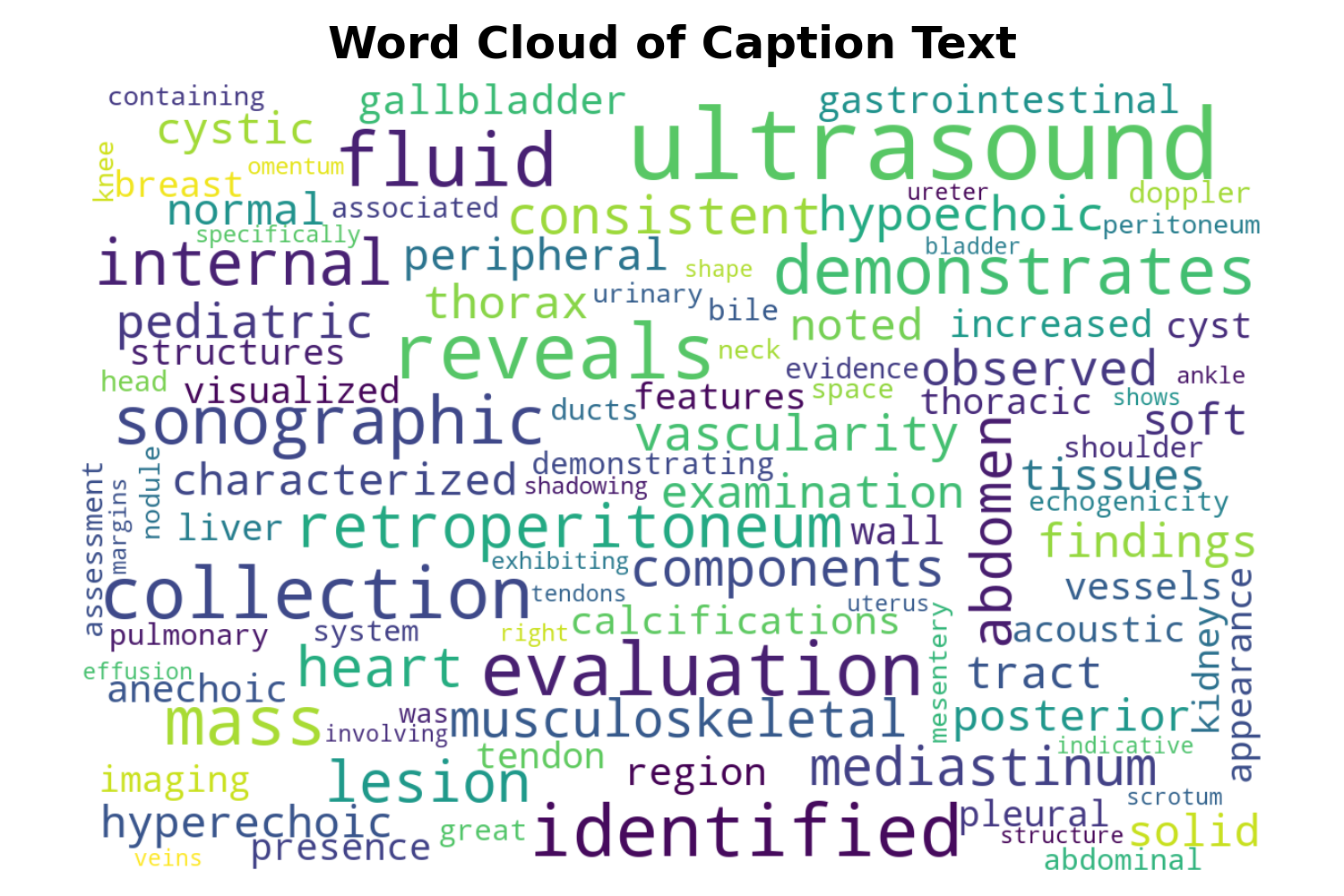} 
    \end{minipage}}
    
    \caption{\textbf{Dataset statistics of the US-365K corpus.} 
    (a) Frequency distribution of the most common diagnostic findings extracted from the curated text. 
    (b) Word cloud summarizing the vocabulary usage in the full caption corpus. 
    (c) Distribution of caption lengths, reflecting the conciseness and variability of clinical reporting.}
    \label{datastats}
    
\end{figure}

To provide a deeper understanding of the composition and linguistic characteristics of US-365K, we present extended statistical analyses in Figure~\ref{datastats}. These visualizations complement the main paper and highlight the dataset's anatomical breadth, diagnostic diversity, and rich textual attributes.

\subsection{Distribution of Diagnostic Findings}

Figure~\ref{datastats}(a) summarizes the frequency distribution of the top 15 diagnostic findings extracted from the refined captions. The histogram exhibits a characteristic long-tailed pattern:

\begin{itemize}
    \item High-frequency findings (e.g., \textit{fluid collection}, \textit{mass}) provide abundant examples of common clinical presentations.
    \item Mid-frequency findings (e.g., \textit{normal appearance}, \textit{cystic components}, \textit{increased vascularity}) represent a balanced mixture of organ- and lesion-level observations.
    \item Low-frequency but clinically relevant entities (e.g., \textit{tubular/linear structures}, \textit{irregular margins}) contribute rare but meaningful diagnostic cues.
\end{itemize}

This distribution confirms that the dataset captures both prevalent pathologies and infrequent but diagnostically important findings, forming a comprehensive representation of real-world sonographic examinations.

\subsection{Caption Vocabulary Analysis}

Figure~\ref{datastats}(b) visualizes the linguistic space of the caption corpus through a word cloud generated from the normalized token distribution. The vocabulary demonstrates the following characteristics:

\begin{itemize}
    \item A balanced mixture of general medical terminology (e.g., ``ultrasound'', ``evaluation'', ``identified'') and specialized sonographic descriptors (e.g., ``hyperechoic'', ``cystic'', ``vascularity'').
    \item High-frequency anatomical terms (e.g., ``gallbladder'', ``retroperitoneal'', ``abdomen'') consistent with the broad organ-level coverage defined by UHAT.
    \item Abundant descriptors of morphology and echogenicity (e.g., ``anechoic'', ``shadowing'', ``components''), reflecting fine-grained imaging semantics.
\end{itemize}

The vocabulary composition highlights both clinical specificity and linguistic diversity, supporting robust training for text-guided ultrasound understanding.

\subsection{Caption Length Distribution}

Figure~\ref{datastats}(c) illustrates the caption length distribution across the dataset. The majority of captions fall within 10–25 words, consistent with the concise nature of radiology reporting. The distribution exhibits:

\begin{itemize}
    \item A mode around 15 words, corresponding to compact diagnostic summaries.
    \item A moderate tail toward longer captions ($>$ 40 words), typically arising from multi-lesion descriptions or more detailed clinical reasoning.
    \item Very few extremely short captions, due to our multi-stage reconstruction process designed to ensure descriptive completeness while avoiding redundancy.
\end{itemize}

This distribution demonstrates that US-365K preserves both the brevity and variability inherent to clinical ultrasound documentation.

\subsection{Summary}

Together, the visualizations in Figure~\ref{datastats} show that US-365K offers:

\begin{itemize}
    \item Anatomical comprehensiveness: 9 body systems and 52 organs;
    \item Diagnostic diversity: coverage spanning common and rare findings;
    \item Linguistic richness: well-structured captions aligned with real-world clinical reporting.
\end{itemize}

These properties collectively make US-365K a strong foundation for ultrasound-focused vision--language pretraining, structured medical understanding, and downstream diagnostic reasoning tasks.

\section{US-365K Dataset Splits}
\label{sec:appendix_splits}

To ensure strict patient-level isolation, robust generalization assessment, and transparent reproducibility, we adopt a case-level split protocol and provide comprehensive corpus statistics. Unless otherwise specified, all analyses and evaluations in the main paper adhere to this protocol.

\subsection{Case-level Split Protocol and Rationale}
\label{subsec:case_split_protocol}
Ultrasound studies are inherently case-centric: multiple frames (including video-derived frames) within a single clinical case may share anatomy, pathology, scanning plane, and narrative context. Random image-wise splitting can therefore induce inadvertent leakage and optimistic bias. To prevent this, we perform splits at the case level, ensuring that no images or captions from the same clinical case appear across different splits (train/validation/test). This design supports faithful evaluation of generalization to unseen cases and mirrors realistic deployment scenarios.

We adopt a 6:2:2 split (training:validation:test) at the case level, stratified over the combined multi-source corpus to preserve source diversity in each partition. Table~\ref{tab:cases_overview} summarizes the split sizes. The final materialized image--text pairs per split are reported in Table~\ref{tab:images_overview}.

\begin{table}[ht]
\centering
\caption{Case-level split overview for US-365K. Splits are mutually exclusive by case.}
\label{tab:cases_overview}
\begin{tabular}{lrr}
\toprule
Partition & \#Cases & Proportion \\\midrule
Train & 7{,}005 & 60.0\% \\
Validation & 2{,}336 & 20.0\% \\
Test & 2{,}335 & 20.0\% \\\midrule
Total & 11{,}676 & 100\% \\
\bottomrule
\end{tabular}
\end{table}

\begin{table}[ht]
\centering
\caption{Image--text pair counts per split for US-365K.}
\label{tab:images_overview}
\begin{tabular}{lrr}
\toprule
Partition & \#Image--Text Pairs & Proportion \\\midrule
Train & 218{,}402 & 59.94\% \\
Validation & 74{,}044 & 20.33\% \\
Test & 71{,}919 & 19.73\% \\\midrule
Total & 364{,}365 & 100\% \\
\bottomrule
\end{tabular}
\end{table}

\subsection{Stratification and Leakage Prevention}
\label{subsec:leakage}
We construct splits by sampling at the case level with the following principles: (1) mutual exclusivity of cases across splits, (2) preservation of per-source composition to first order, and (3) invariance to within-case frame multiplicity (video-derived frames remain within the case they originate from). This prevents cross-split near-duplicates and leakage through subfigures, multi-frame snippets, or shared narratives.

To further reduce subtle leakage risks in composite figures (e.g., multi-panel images with shared captions), we associate all subfigures and subcaptions to the same parent case and enforce split-consistency at the parent level. The same policy applies to video-derived frames.


\section{Prompts Design and Templates}
\label{prompts}
We release the exact prompt texts used in all stages. To fit the two-column layout, we provide (1) short context bullets, (2) the full template in a compact code block, and (3) JSON schemas as inline lists. Placeholders in braces are programmatically injected.

\subsection{Caption Reconstruction Prompts}
This pipeline has three templates: Stage 1 (Anatomy tags), Stage 2 (Lesion attributes), and (3) Tag-to-Caption generation. All prompts enforce closed-world extraction, taxonomy-locked labels, and traceable rationale.

\subsubsection{Stage 1: Anatomical Tag Extraction Prompts}

Goal: Extract UHAT-aligned two-level anatomy (Body system, Organ).

\noindent Placeholders: $\{classification\_tree\}$, $\{output\_format\}$, $\{description\}$.

\noindent Output JSON schema:
$Anatomy\_Body\_system\_level$, $Anatomy\_Organ\_level$, $Explanation: string.$ 

\begin{figure}[th]
    \centering
    \includegraphics[width = 1\linewidth]{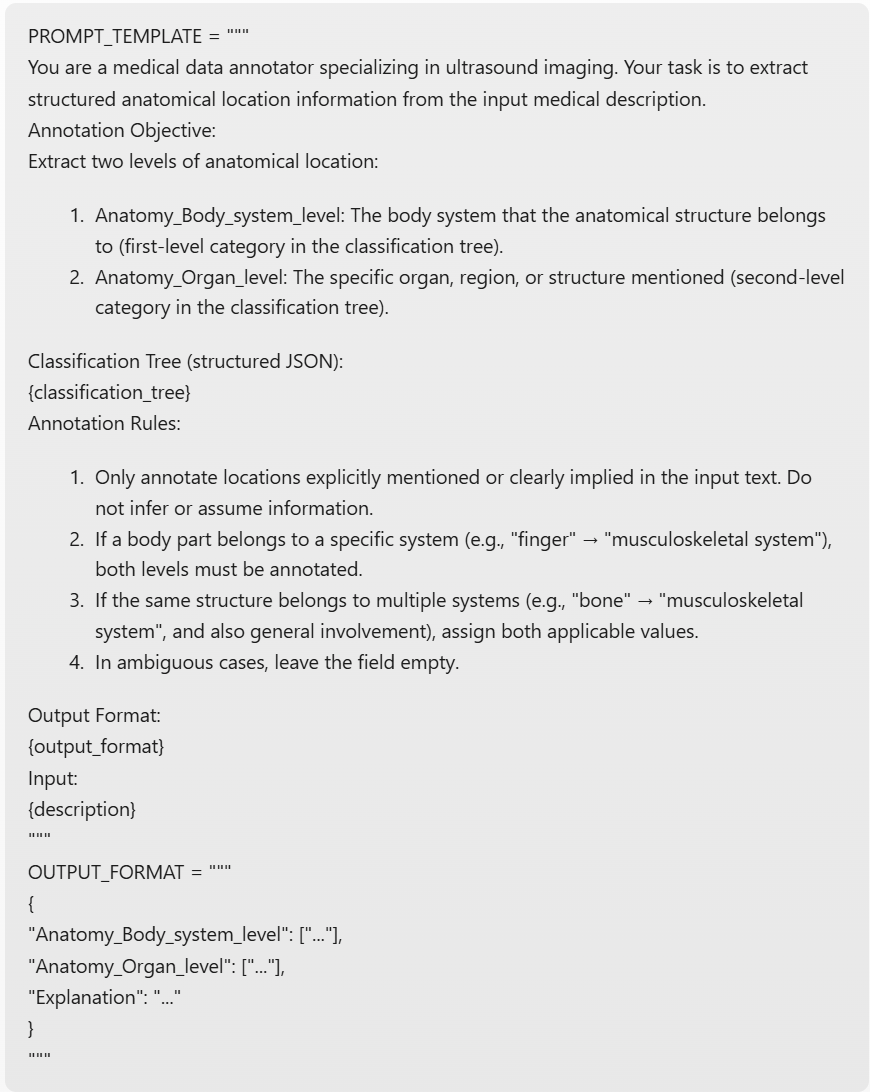}
    \caption{Prompts for Stage 1: Anatomical Tag Extraction}
    \label{prompt1}
\end{figure}

\noindent The complete prompt template is visualized in Figure~\ref{prompt1}.

\subsubsection{Stage 2: Lesion Attribute Extraction Prompts}
Goal: Extract UDAF-aligned attributes across remaining seven dimensions.

\noindent Placeholders: \{$attribute$\}, \{$output\_format$\}, \{$description$\}.

\noindent Rule: The model is instructed to extract only explicitly stated or clearly implied information. All labels must be from the closed sets. 

\begin{figure}[th]
    \centering
    \includegraphics[width = 1\linewidth]{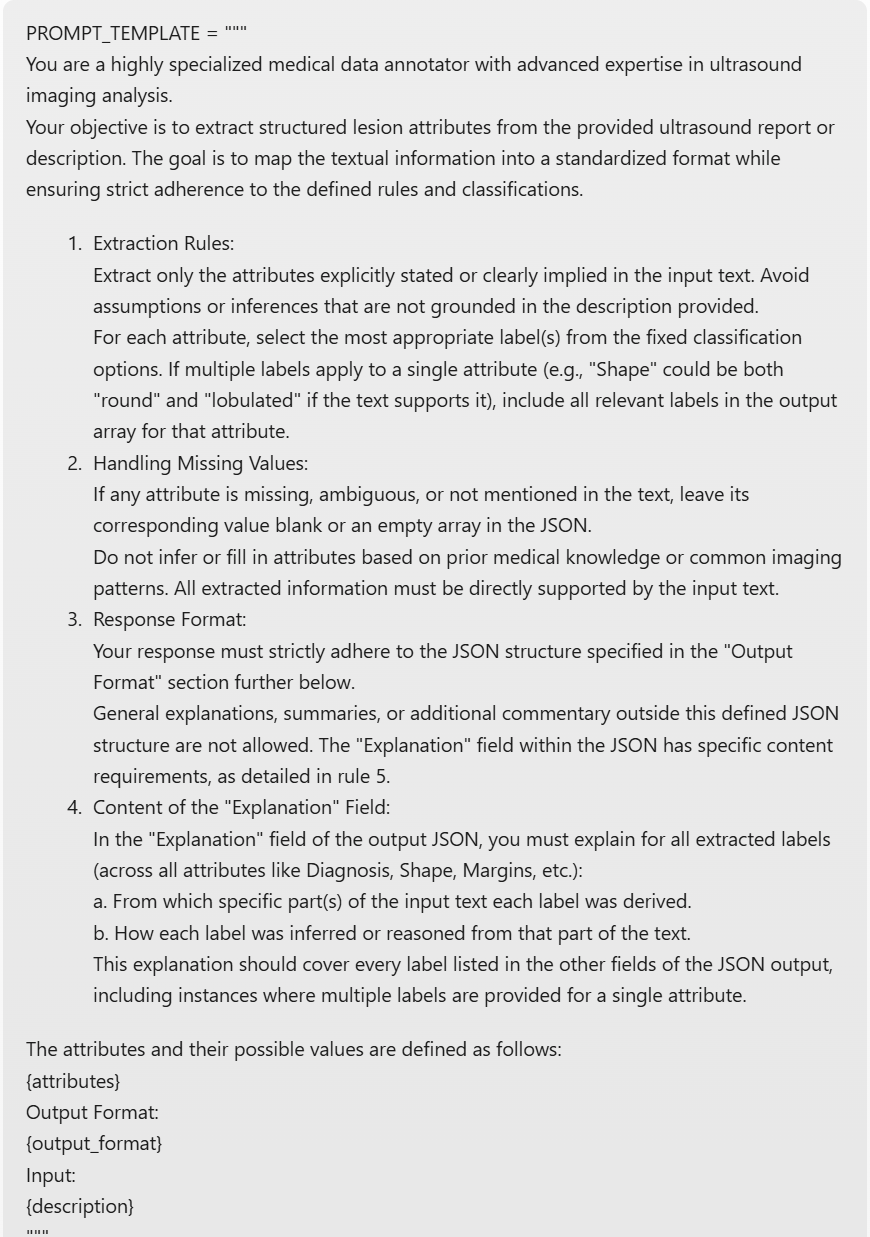} 
    \includegraphics[width = 1\linewidth]{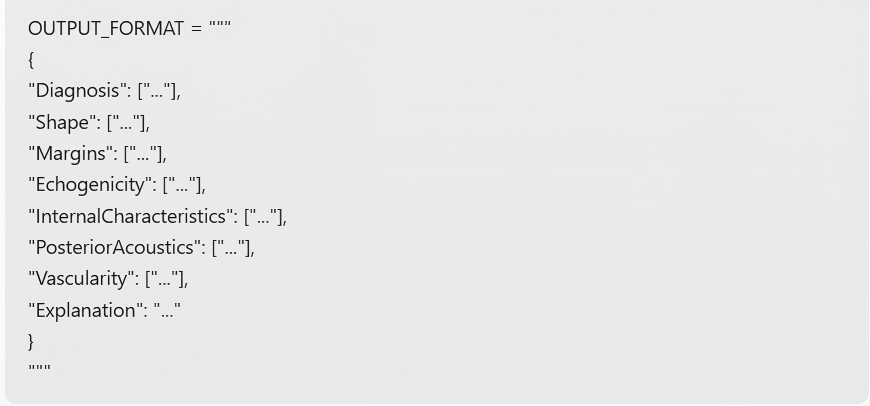} 
    \caption{Prompts for Stage 2: Lesion Attribute Extraction}
    \label{prompt2}
\end{figure}

\noindent The complete prompt template is visualized in Figure~\ref{prompt2}.

\subsubsection{Stage 3: Tag-to-Caption Generation Prompts}
Goal: Generate three clinically faithful sentences from tags, strictly adhering to the input features to prevent hallucination.

\noindent Placeholders: $\{Body\_system\_level\}$, $\{Organ\_level\}$, $\{diagnosis\}$, $\{shape\}$, $\{margin\}$, $\{echogenicity\}$, $\{internal\_features\}$, $\{posterior\_features\}$, $\{vascularity\}$, $\{Explanation\}.$

\begin{figure}[th]
    \centering
    \includegraphics[width = 1\linewidth]{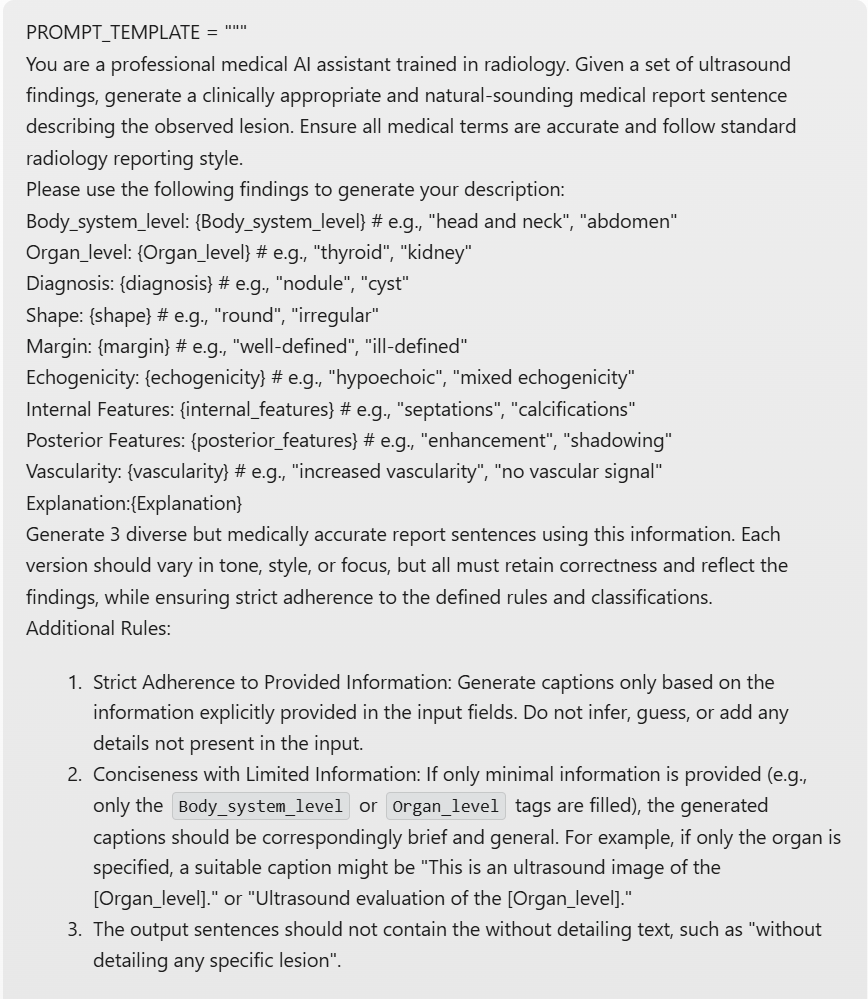} 
    \includegraphics[width = 1\linewidth]{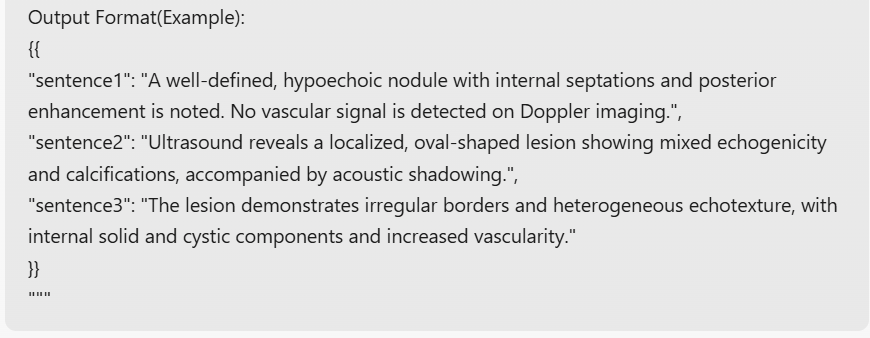} 
    \caption{Prompts for Stage 3: Tag-to-Caption Generation}
    \label{prompt3}
\end{figure}

\noindent In scenarios where only anatomical tags (system/organ) are present, the model is instructed to generate brief, generic descriptive statements. The full prompt template is visualized in Figure~\ref{prompt3}.

\subsection{Multi-task Ultrasound Classification Prompts}
We instantiate CLIP-style text prompts per task with concise, taxonomy-aligned descriptions. The mapping below defines the exact text strings used during evaluation. For space, we provide the full dictionary as a compact listing; each key is a class label, and the value is the prompt text.

\begin{itemize}

\item \textbf{``Task 1'':} \{\\
    ``Abdomen and retroperitoneum'': ``a ultrasound image of Abdomen and retroperitoneum'', \\
    ``Urinary Tract and male reproductive system'': ``a ultrasound image of Urinary Tract and male reproductive system'', \\
    ``Gynaecology'': ``a ultrasound image of Gynaecology'', \\
    ``Head and Neck'': ``a ultrasound image of Head and Neck'', \\
    ``Breast and Axilla'': ``a ultrasound image of Breast and Axilla'', \\
    ``Musculoskeletal Joints and Tendons'': ``a ultrasound image of Musculoskeletal Joints and Tendons'', \\
    ``Thorax'': ``a ultrasound image of Thorax'', \\
    ``Pediatrics'': ``a ultrasound image of Pediatrics'', \\
    ``Peripheral vessels'': ``a ultrasound image of Peripheral vessels''
\}

\item \textbf{``Task 2'':} \{\\
    ``Liver'': ``a ultrasound image of Liver'', \\
    ``Gallbladder and bile ducts'': ``a ultrasound image of Gallbladder and bile ducts'', \\
    ``Pancreas'': ``a ultrasound image of Pancreas'', \\
    ``Spleen'': ``a ultrasound image of Spleen'', \\
    ``Appendix'': ``a ultrasound image of Appendix'', \\
    ``Gastrointestinal tract'': ``a ultrasound image of Gastrointestinal tract'', \\
    ``Peritoneum mesentery and omentum'': ``a ultrasound image of Peritoneum mesentery and omentum'', \\
    ``Retroperitoneum and great vessels'': ``a ultrasound image of Retroperitoneum and great vessels'', \\
    ``Adrenal glands'': ``a ultrasound image of Adrenal glands'', \\
    ``Abdominal wall'': ``a ultrasound image of Abdominal wall'', \\
    ``Kidney and ureter'': ``a ultrasound image of Kidney and ureter'', \\
    ``Bladder'': ``a ultrasound image of Bladder'', \\
    ``Scrotum'': ``a ultrasound image of Scrotum'', \\
    ``Penis and perineum'': ``a ultrasound image of Penis and perineum'', \\
    ``Uterus'': ``a ultrasound image of Uterus'', \\
    ``Adnexa'': ``a ultrasound image of Adnexa'', \\
    ``Vagina'': ``a ultrasound image of Vagina'', \\
    ``Thyroid gland'': ``a ultrasound image of Thyroid gland'', \\
    ``Parathyroid glands'': ``a ultrasound image of Parathyroid glands'', \\
    ``Salivary glands'': ``a ultrasound image of Salivary glands'', \\
    ``Lymph nodes'': ``a ultrasound image of Lymph nodes'', \\
    ``Ocular'': ``a ultrasound image of Ocular'', \\
    ``Ear'': ``a ultrasound image of Ear'', \\
    ``Larynx'': ``a ultrasound image of Larynx'', \\
    ``Breast'': ``a ultrasound image of Breast'', \\
    ``Axilla'': ``a ultrasound image of Axilla'', \\
    ``Shoulder'': ``a ultrasound image of Shoulder'', \\
    ``Elbow'': ``a ultrasound image of Elbow'', \\
    ``Wrist and carpus'': ``a ultrasound image of Wrist and carpus'', \\
    ``Fingers'': ``a ultrasound image of Fingers'', \\
    ``Hip groin and buttock'': ``a ultrasound image of Hip groin and buttock'', \\
    ``Knee'': ``a ultrasound image of Knee'', \\
    ``Ankle'': ``a ultrasound image of Ankle'', \\
    ``Foot'': ``a ultrasound image of Foot'', \\
    ``Peripheral nerves'': ``a ultrasound image of Peripheral nerves'', \\
    ``Soft tissues'': ``a ultrasound image of Soft tissues'', \\
    ``Skull'': ``a ultrasound image of Skull'', \\
    ``Pulmonary'': ``a ultrasound image of Pulmonary'', \\
    ``Pleural space'': ``a ultrasound image of Pleural space'', \\
    ``Heart and mediastinum'': ``a ultrasound image of Heart and mediastinum'', \\
    ``Thoracic wall'': ``a ultrasound image of Thoracic wall'', \\
    ``Pediatric abdomen and retroperitoneum'': ``a ultrasound image of Pediatric abdomen and retroperitoneum'', \\
    ``Pediatric urinary tract'': ``a ultrasound image of Pediatric urinary tract'', \\
    ``Pediatric scrotum'': ``a ultrasound image of Pediatric scrotum'', \\
    ``Pediatric gynaecological pathology and infant breast'': ``a ultrasound image of Pediatric gynaecological pathology and infant breast'', \\
    ``Pediatric head and neck'': ``a ultrasound image of Pediatric head and neck'', \\
    ``Neonatal brain and spine'': ``a ultrasound image of Neonatal brain and spine'', \\
    ``Infant hip and knee'': ``a ultrasound image of Infant hip and knee'', \\
    ``Pediatric thorax'': ``a ultrasound image of Pediatric thorax'', \\
    ``Peripheral arteries'': ``a ultrasound image of Peripheral arteries'', \\
    ``Peripheral veins'': ``a ultrasound image of Peripheral veins'', \\
    ``Dialysis fistula'': ``a ultrasound image of Dialysis fistula''
\}
\item \textbf{``Task 3'':} \{\\
    ``nodule'': ``a nodule in an ultrasound image'', \\
    ``cyst'': ``a cyst in an ultrasound image'', \\
    ``mass'': ``a mass in an ultrasound image'', \\
    ``fluid collection'': ``a fluid collection in an ultrasound image'', \\
    ``normal appearance'': ``normal appearance in an ultrasound image''
\}

\item \textbf{``Task 4'':} \{\\
    ``round'': ``a round lesion in an ultrasound image'', \\
    ``oval'': ``an oval lesion in an ultrasound image'', \\
    ``lobulated'': ``a lobulated lesion in an ultrasound image'', \\
    ``tubular/linear'': ``a tubular or linear lesion in an ultrasound image'', \\
    ``nodular'': ``a nodular lesion in an ultrasound image'', \\
    ``flattened'': ``a flattened lesion in an ultrasound image'', \\
    ``irregular'': ``an irregular lesion in an ultrasound image''
\}

\item \textbf{``Task 5'':} \{\\
    ``well-defined'': ``a lesion with well-defined margins in an ultrasound image'', \\
    ``ill-defined/indistinct'': ``a lesion with ill-defined/indistinct margins in an ultrasound image''
\}

\item \textbf{``Task 6'':} \{\\
    ``anechoic'': ``an anechoic lesion in an ultrasound image'', \\
    ``hypoechoic'': ``a hypoechoic lesion in an ultrasound image'', \\
    ``isoechoic'': ``an isoechoic lesion in an ultrasound image'', \\
    ``hyperechoic'': ``a hyperechoic lesion in an ultrasound image'', \\
    ``mixed echogenicity'': ``a lesion with mixed echogenicity in an ultrasound image''
\}

\item \textbf{``Task 7'':} \{\\
    ``cystic components'': ``a lesion with cystic components in an ultrasound image'', \\
    ``calcifications'': ``a lesion with calcifications in an ultrasound image'', \\
    ``septations'': ``a lesion with septations in an ultrasound image'', \\
    ``solid components'': ``a lesion with solid components in an ultrasound image'', \\
    ``mixed cystic and solid mass'': ``a mixed cystic and solid mass in an ultrasound image''
\}

\item \textbf{``Task 8'':} \{\\
    ``enhancement'': ``a lesion with posterior acoustic enhancement in an ultrasound image'', \\
    ``shadowing'': ``a lesion with posterior acoustic shadowing in an ultrasound image''
\}

\item \textbf{``Task 9'':} \{\\
    ``reduced/diminished vascularity'': ``a lesion with reduced or diminished vascularity in an ultrasound image'', \\
    ``normal/regular vascularity'': ``a lesion with normal or regular vascularity in an ultrasound image'', \\
    ``no vascularity'': ``a lesion with no vascularity in an ultrasound image'', \\
    ``increased vascularity'': ``a lesion with increased vascularity in an ultrasound image'', \\
    ``indeterminate/inhomogeneous vascularity'': ``a lesion with inhomogeneous or indeterminate vascularity in an ultrasound image''
\}
\end{itemize}

\subsection{Downstream Classification Prompts}
For each dataset, we use a single-sentence, class-conditioned template to form zero-shot labels. In the templates below, the placeholder \texttt{\{class\}} is dynamically replaced with the specific category name of the dataset.

\begin{itemize}
\item BUSBRA: “a breast ultrasound image showing \texttt{\{class\}} lesion”
\item GIST514-DB: “an endoscopic image showing \texttt{\{class\}}”
\item BreastMNIST: “a pathology image showing \texttt{\{class\}} breast lesion”
\item Breast: “a breast ultrasound image showing \texttt{\{class\}}”
\end{itemize}

\section{Experiment Results}

\subsection{Downstream Tasks}

\begin{table*}[!htp]
\centering
\setlength{\tabcolsep}{4pt}
\caption{\textbf{Downstream task generalization comparison across 4 ultrasound datasets.} We report the accuracy performance under ZS, LP, and FT settings. * Models cannot be retrained due to unavailable code.}
\label{tab:acc_comparison}
\vspace{-0.1in}
\footnotesize
\begin{tabular}{l|ccc|ccc|ccc|ccc|ccc}
\toprule
\textbf{Method} & \multicolumn{3}{c|}{\textbf{BUS-BRA}} & \multicolumn{3}{c|}{\textbf{GIST514-DB}} & \multicolumn{3}{c|}{\textbf{BreastMNIST}} & \multicolumn{3}{c|}{\textbf{Breast}} & \multicolumn{3}{c}{\textbf{Average Rank}} \\
 & ZS & LP & FT & ZS & LP & FT & ZS & LP & FT & ZS & LP & FT & ZS & LP & FT \\
\midrule
\multicolumn{16}{l}{\textit{General CLIP}} \\
\midrule
CLIP & 51.49 & 70.16 & 86.15 & 49.03 & 62.58 & 76.13 & 28.33 & 73.92 & 90.38 & 50.77 & 69.66 & 84.19 & 6.25 & 6.5 & 3.75\\
SigLIP* & 61.12 & 69.80 & 85.61 & 48.44 & 67.10 & \textbf{81.40} & 51.76 & 78.85 & 86.54 & 54.74 & 74.36 & 84.62 & 4.5 & 5 & 4\\
MetaCLIP & \textbf{65.97} & 67.85 & 84.19 & 48.05 & 58.06 & 73.55 & 41.37 & 73.72 & \textbf{91.67} & 71.97 & 67.52 & 84.62 & 4.25 & 8 & 4.25 \\
\midrule
\multicolumn{16}{l}{\textit{Medical CLIP}} \\
\midrule
PMC-CLIP & 35.25 & 75.31 & 82.06 & 50.10 & 63.87 & 68.39 & \textbf{69.87} & 79.49 & 87.82 & 29.36 & 72.22 & 81.20  & 5 & 4.75 & 7.5\\
MedCLIP* & 36.21 & 71.94 & \textbf{87.21} & 51.36 & 68.39 & 72.26 & 28.44 & 78.85 & 86.54 & \textbf{74.49} & 67.95 & 89.32 & 4.25 & 4.75 & 4.25\\
UniMed-CLIP & 65.14 & 77.44 & 84.10 & 48.52 & 65.81 & 74.84 & 66.28 & 86.54 & 91.03 & 68.42 & 76.92 & 89.32  & 3.5 & 3 & 3.75 \\
BiomedCLIP & 33.33 & 77.44 & 84.19 & 51.56 &\textbf{ 69.03} & 74.84 & 47.69 & 84.62 & 89.10 & 52.44 & \textbf{79.49} & 88.89 & 5.25 & 1.75 & 4.25 \\
\textbf{Ultrasound-CLIP} & 54.93 & \textbf{78.86} & 84.55 & \textbf{53.89} & 68.39 & 72.90 & 49.10 & \textbf{88.46} & \textbf{92.95} & 70.26 & 77.35 & \textbf{91.45}  & \textbf{3} & \textbf{1.5} & \textbf{3}\\
\bottomrule
\end{tabular}
\vspace{-0.1in}
\end{table*}
The results shown in Table ~\ref{tab:acc_comparison} validate the impact of pre-training strategies on cross-dataset transferability. Overall, general-purpose CLIP models exhibit stable performance in the zero-shot setting, yet a noticeable semantic domain gap remains when applied to medical ultrasound. In contrast, most medical-specific CLIP variants achieve stronger discriminative power under limited-parameter learning and full fine-tuning, demonstrating the importance of medical text–image priors.

Among all compared methods, Ultrasound-CLIP consistently shows more balanced and robust generalization across datasets and evaluation protocols. In the zero-shot setting, it achieves the highest accuracy on GIST514-DB, reaching 53.89\%, respectively, indicating that meaningful alignment between ultrasound structures and lesion semantics can be obtained without downstream training. In the LP setting, Ultrasound-CLIP attains the best or near-best performance on three datasets and achieves the top average ranking of 1.5, highlighting its suitability for low-annotation regimes. Under full fine-tuning, the model remains competitive, obtaining the highest accuracy on both BreastMNIST and Breast, and reaching the overall best average ranking.

Although several medical CLIP models outperform others in isolated tasks, Ultrasound-CLIP demonstrates more consistent cross-domain robustness across the four datasets and three evaluation settings. This stability stems from its large-scale pre-training corpus curated specifically for ultrasound understanding, enabling superior representation of anatomical structures, acoustic patterns, and lesion morphology compared to general vision–language models and existing medical multimodal models. Overall, the results indicate that Ultrasound-CLIP provides a reliable foundation model with strong transferability for a wide range of real-world ultrasound intelligence applications.

\subsection{Efficiency Analysis}
The results shown in Table~\ref{tab:efficiency_analysis}. Ultrasound-CLIP (176.28M Params) is almost 5×smaller than SigLIP (877.96M), indicating that performance gains are not due to model scale. Notably, Ultrasound-CLIP achieves 74.97 FPS, surpassing all baselines, validating its feasibility for real-time deployment.

\begin{table*}[!htp]
    \centering
    \caption{\textbf{Efficiency analysis.} We report the efficiency analysis results.}
    \vspace{-0.1in}
    \footnotesize
    \begin{tabular}{lcccccccc}
        \toprule
        \multirow{2}{*}{Model} 
        & \multirow{2}{*}{CLIP} 
        & \multirow{2}{*}{SigLIP} 
        & PMC 
        & UniMed 
        & Med 
        & Meta 
        & Biomed 
        & Ultrasound \\
         &  &  & CLIP & CLIP & CLIP & CLIP & CLIP & CLIP \\
        \midrule
        Params (M) & 102 & 878 & 200 & 196 & 133 & 151 & 196 & \textbf{176} \\
        FLOPs (G)  & 7.36 & 326.7 & 13.5 & 33.0 & 4.73 & 4.89 & 25.0 & 26.9 \\
        FPS        & 44.4 & 16.8 & 36.1 & 45.7 & 33.9 & 32.6 & 43.5 & \textbf{75.0} \\
        \bottomrule
    \end{tabular}
    \label{tab:efficiency_analysis}
    \vspace{-0.1in}
\end{table*}


\subsection{Robustness to Sparse Captions.}
We group samples by caption sparsity (Table~\ref{tab:Robustness}). The model maintains robust accuracy with different textual supervision, confirming its applicability to real-world clinical data.
\begin{table*}[!htp]
    \caption{\textbf{Robustness to sparse captions.}}
    \vspace{-0.1in}
    \label{tab:Robustness}
    \footnotesize
    \centering
    \begin{tabular}{@{}lccccccccc@{}}
    \toprule
    \textbf{\#Tasks}   & 1 & 2 & 3 & 4 & 5 & 6 & 7 & 8 & 9 \\
    \midrule
    \textbf{\#Samples} & 10 & 25972 & 26094 & 13145 & 5872 & 1485 & 225 & 6 & 2 \\
    \textbf{Acc (\%)}  & 80.00 & 45.72 & 59.40 & 54.30 & 45.61 & 50.25 & 48.19 & 37.50 & 61.11 \\
    \bottomrule
    \end{tabular}
    \vspace{-0.1in}
\end{table*}

\subsection{Case Studies}

\begin{figure}[th]
    \centering
    \includegraphics[width = 1\linewidth]{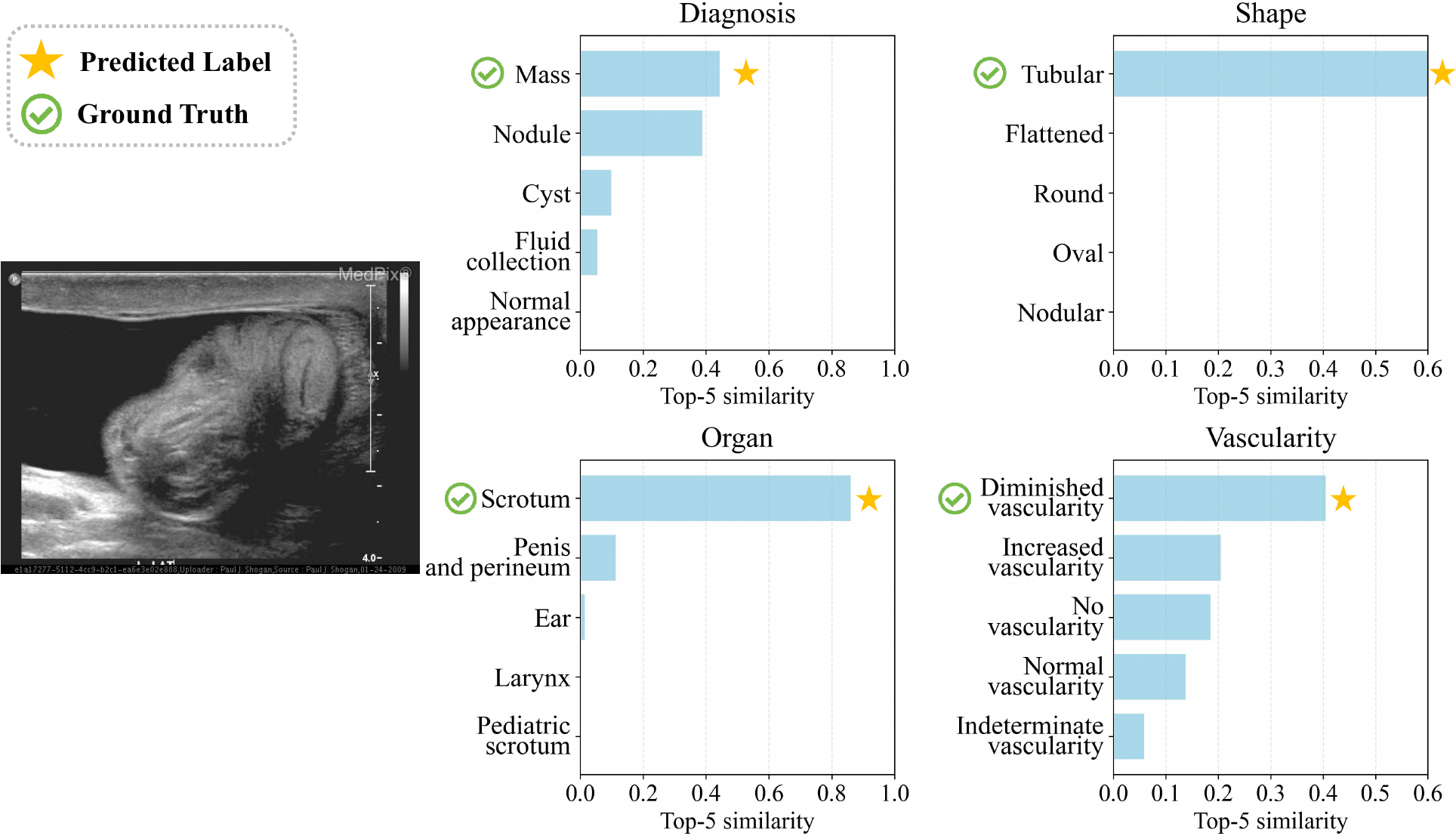}
    \caption{Probabilistic alignment in unstructured diagnosis.}
    \label{fig:casestudy1}
    \vspace{-0.1in}
\end{figure}

\begin{figure}[th]
    \centering
    \includegraphics[width = 1\linewidth]{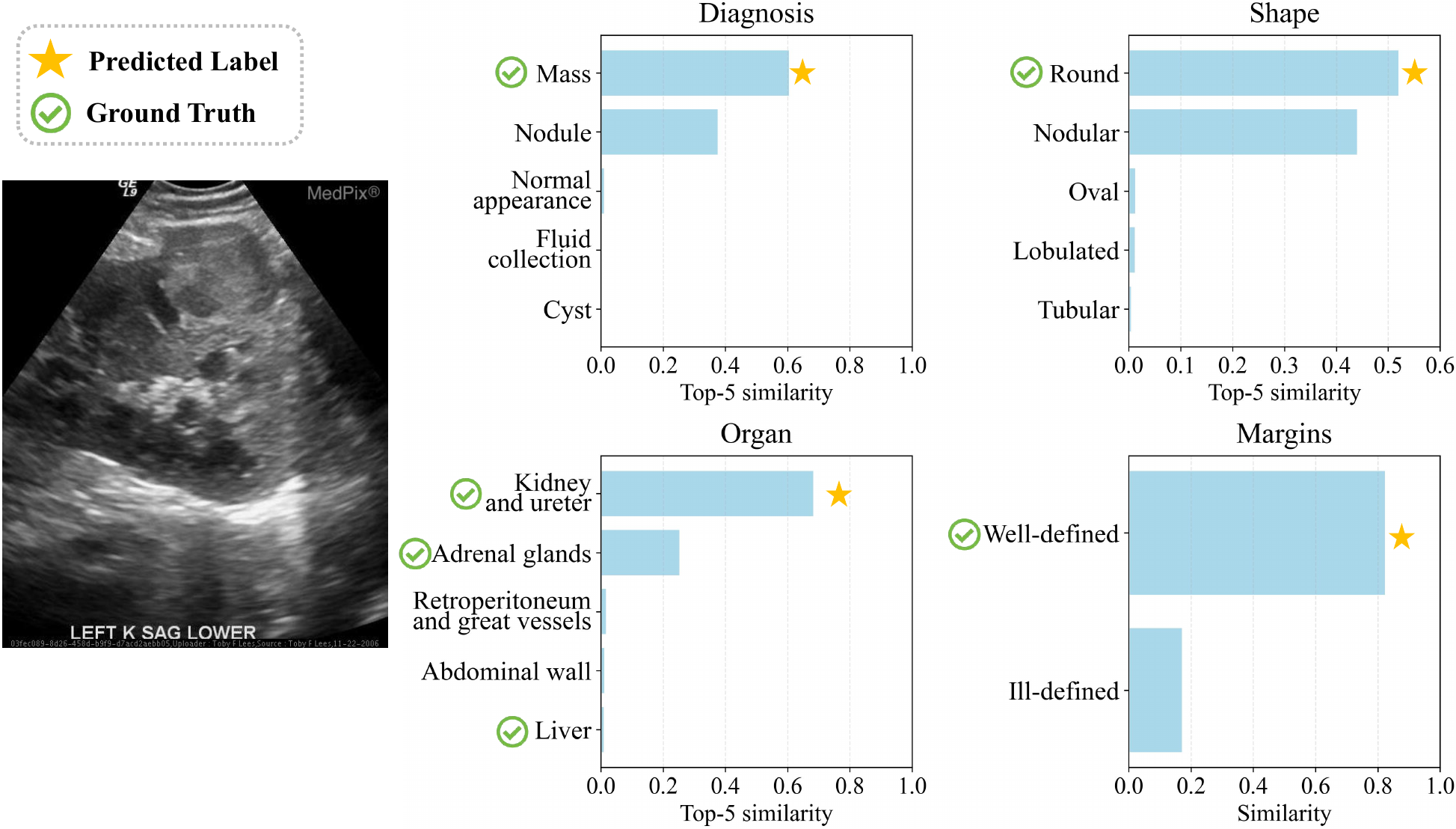}
    \caption{Multi-label clinical insight.}
    \label{fig:casestudy2}
    \vspace{-0.1in}
\end{figure}

\noindent\textbf{Case Study: Probabilistic Alignment in Unstructured Diagnosis.} 
Figure~\ref{fig:casestudy1} illustrates a diagnostic case where the ground truth involves ambiguity between the labels ``Mass'' and ``Fluid collection''. Ultrasound-CLIP correctly identifies ``Mass'' as the primary label, aligning with the radiological description’s emphasis on anatomical and shape attributes. However, what sets the model apart is its ability to capture the probabilistic proximity of alternative labels. Specifically, ``Fluid collection,'' although not predicted as the top label, is ranked second, with nearly comparable confidence. This highlights the model’s capacity for nuanced reasoning within diagnostic uncertainty.

Additionally, supporting diagnostic attributes predicted by the model reinforce its alignment with the clinical context. For instance, ``Rounded'' for shape and ``Well-defined'' for margins are consistent with ``Mass'' but can also correspond to the fluid-filled presentation of certain cystic lesions. By effectively balancing the structured reasoning of primary and secondary outputs, Ultrasound-CLIP exhibits interpretability and robustness that mirror human diagnostic reasoning under conflicting or partially ambiguous cases. These results demonstrate the model’s probabilistic reasoning depth and suggest its applicability across diverse clinical scenarios, even where ground truth spans multiple diagnostic possibilities.

\noindent\textbf{Case Study: Multi-label Clinical Insight.} In Figure~\ref{fig:casestudy2}, we present a case that showcases the adaptability of Ultrasound-CLIP to a complex diagnostic scenario involving multi-label clinical annotations. The ground truth includes multiple valid diagnoses, notably ``Nodule'' and ``Cyst'', reflecting the heterogeneous nature of this lesion’s presentation. While our model is designed to predict a single primary diagnosis, its ranked predictions for supporting labels demonstrate high clinical alignment. Specifically, ``Nodule'' was identified as the primary label with a top confidence score, matching the main diagnosis in the report. Notably, ``Cyst'' while not the top-ranked prediction, is captured within the higher-ranked outputs. This demonstrates that the model’s probabilistic reasoning effectively maintains relevance across alternative diagnostic interpretations.

More critically, the supporting attributes predicted for ``Nodule'' (e.g., ``Well-defined'' for margins and ``Hypoechoic'' for echogenicity) are also highly consistent with the characteristics of ``Cyst'', indicating a nuanced understanding of overlapping diagnostic features. This case exemplifies Ultrasound-CLIP’s ability not only to prioritize a single primary label but also to implicitly reflect the multi-label complexity inherent in clinical practice. The results further underscore the framework’s capability to capture structured relationships between diagnostic attributes and provide clinically coherent reasoning, even in single-label prediction tasks.

%

\end{document}